\documentclass[11pt]{article}

\PassOptionsToPackage{table,xcdraw}{xcolor}
\usepackage[final]{acl}

\usepackage{times}
\usepackage{latexsym}
\usepackage[T1]{fontenc}
\usepackage[utf8]{inputenc}
\usepackage{microtype}
\usepackage{inconsolata}
\usepackage{graphicx}

\usepackage{amsmath}
\usepackage{booktabs}
\usepackage{multirow}
\usepackage{ulem}
\usepackage{array}
\usepackage{tabularx}
\usepackage{xspace}
\usepackage{url}
\usepackage{pifont}
\usepackage{tikz}
\usetikzlibrary{arrows.meta,positioning,shapes.geometric}
\usepackage{rotating}

\newcommand{\sys}[1]{\textsc{#1}}
\newcommand{\averitec}{AVeriTeC}
\newcommand{\climatefever}{ClimateFEVER}
\newcommand{\climatecheck}{ClimateCheck}
\newcommand{\scifact}{SciFact}
\newcommand{\ms}[2]{#1{\,\scriptsize$\pm$\,#2}}

\definecolor{darkgreen}{rgb}{0.0, 0.5, 0.0}
\definecolor{verylightgray}{rgb}{0.97, 0.97, 0.97}
\newcommand{\cmark}{\textcolor{darkgreen}{{\ding{51}}}} 
\newcommand{\pmark}{\textcolor{orange}{{\ding{115}}}} 

\title{How Robust Are Automated Fact-Checking Systems? \\ A Cross-Benchmark Evaluation}

\author{
 \textbf{Aida Usmanova\textsuperscript{1}},
 \textbf{Zangir Iklassov\textsuperscript{2}},
 \textbf{Markus Leippold\textsuperscript{3}},
 \textbf{Ricardo Usbeck\textsuperscript{1}}
\\
\\
 \textsuperscript{1}Leuphana University of Lüneburg,
 \textsuperscript{2}MBZUAI,
 \textsuperscript{3}University of Zurich
\\
 \small{
   \textbf{Correspondence:} \href{aida.usmanova@stud.leuphana.de}{aida.usmanova@stud.leuphana.de}
 }
}

\begin{document}
\maketitle

\begin{abstract}
Automated fact-checking (AFC) systems retrieve evidence and predict claim veracity, yet evaluations omit simple baselines, systems are developed for a single benchmark and cannot be trusted to generalise across domains.
No prior work cross-evaluates the full two-stage retrieve-then-verify pipeline across diverse datasets,
complementing retrieval-only studies~\cite{thakur2021beir} and single-stage benchmarking studies~\cite{calamai2025benchmarking}.
We benchmark nine models, ranging from random and sparse baselines to fine-tuned transformers,
zero-shot LLMs, and the two highest-ranked systems from the \averitec{}~2025 shared task, across four datasets spanning scientific, open-web, and climate domains.
Three findings stand out:
(1)~on \climatecheck{} claim-only and fine-tuned models outperform zero-shot LLM and top-performing \averitec{} 2025 systems, highlighting that noisy evidence can degrade veracity prediction;
(2)~system rankings are strongly domain- and metric-dependent: the best model on \scifact{} (macro-F1 0.70) drops to 0.31 on \climatecheck{}, while the \averitec{}~2025 winner and runner-up swap rankings based on evaluation metrics and datasets;
(3)~replacing retrieved evidence with gold annotations improves veracity accuracy by 14-22 points across models, confirming retrieval remains primary bottleneck.
We release code, pre-processed datasets, and all results to support reproducible AFC research.%
\footnote{\url{https://github.com/aidausmanova/FCBench}}
\end{abstract}

\section{Introduction}
\label{sec:introduction}

The proliferation of misinformation across news, social media, and scientific discourse has motivated extensive research in automated
fact-checking (AFC)~\cite{thorne-vlachos-2018-automated,guo2022survey}.
Formalised by \citet{vlachos-riedel-2014-fact}, the dominant paradigm established by FEVER shared task~\cite{fever} is a two-stage pipeline (Figure~\ref{fig:pipeline}): a retrieval module selects relevant evidence, and a veracity model predicts whether it \emph{supports}, \emph{refutes}, or is insufficient to judge the claim (\emph{not enough information}, NEI). Inspired by FEVER~\cite{fever}, progress has been made on benchmarks spanning news claims~\cite{wang-2017-liar},
multi-domain evidence~\cite{augenstein-etal-2019-multifc}, multi-hop reasoning~\cite{yang-etal-2018-hotpotqa,jiang-etal-2020-hover},
scientific claims~\cite{scifact,saakyan-etal-2021-covid}, and knowledge-intensive tasks~\cite{petroni-etal-2021-kilt}. Yet the ultimate test of an AFC system is not benchmark rank, but robustness to domain shifts.

AFC systems are developed to combat real-world misinformation, yet we do not know whether reported progress shows genuinely generalisable capabilities. First, systems in shared tasks involve multi-step pipelines and are rarely compared against simple baselines such as TF-IDF retrieval~\cite{manning2008introduction} combined with
logistic regression. Without these lower bounds, it is impossible to tell whether reported gains reflect actual improvements in language understanding or dataset-specific engineering.
Second, misinformation spreads across multiple domains, and reliable AFC systems should operate across diverse settings. However, SOTA systems are usually trained and tested on a single benchmark and may poorly translate to other domains~\cite{guo2022survey}. A system that does not generalise well across domains cannot be trusted in real-world deployment, where domain of incoming claims is unknown in advance.

\citet{calamai2025benchmarking} found that TF-IDF achieves within 5\% of fine-tuned transformers across 29 climate-related NLP benchmarks, and that most datasets contain annotation issues that further obscure real model differences. \citet{thakur2021beir} showed that BM25~\cite{Robertson2009bm25} outperforms neural retrieval
on the majority of 18 out-of-domain information retrieval benchmarks.
However, both studies evaluate only a single pipeline stage in isolation, namely information retrieval or single-task classification.
No prior work cross-evaluates the complete retrieve-then-verify pipeline across structurally diverse domains.

\textbf{We present a unified, transferable evaluation of fact-checking systems} across
four structurally diverse datasets (Figure~\ref{fig:taxonomy}) evaluated under identical
conditions, covering sparse retrieval, fine-tuned transformers, zero-shot LLMs, and
the two highest-ranked systems from the 2025 \averitec{} shared task~\cite{fever2025}, an annual challenge for open-domain AFC in which \sys{AIC~CTU} (the winner) and \sys{Sanctuary} (runner-up) represent strong evidence-based verification systems.
Our study presents three key findings:
\begin{itemize}
  \item \textbf{Competitiveness of classical baselines.}
  Under the evaluated pipeline, logistic regression over sparse retrieval outperformed evidence-conditioned LLMs and shared task top-performers, due to harmful evidence retrieval.
  \item \textbf{Retrieval quality is a main challenge.}
    Replacing retrieved evidence with gold annotations improves veracity accuracy by 14-22 points across models, diminishing gains from switching to stronger veracity models.
  \item \textbf{Rankings are domain- and metric-dependent.}
    No single system dominates across all four benchmarks.
    The best system on \scifact{} (macro-F1 0.700) drops to 0.315 on \climatecheck{}. The \averitec{}~2025 winner \sys{AIC CTU} swaps position with runner-up \sys{Sanctuary} depending on evaluation metrics and datasets, making single-benchmark evaluation an unreliable proxy for general capability.
\end{itemize}

\section{Related Work}
\label{sec:related}

\paragraph{Automated fact-checking.}
Veracity prediction is framed as natural language inference (NLI) over retrieved evidence, a formulation that traces back to SNLI~\cite{bowman-etal-2015-large} and
MultiNLI~\cite{williams-etal-2018-broad}, and was operationalised for AFC by BERT-based cross-encoders~\cite{devlin-etal-2019-bert}.
Graph-based evidence aggregation~\cite{zhou-etal-2019-gear,nie2019combining},
explanation-generating models~\cite{atanasova-etal-2020-generating}, and
contrastive training for robustness~\cite{schuster-etal-2021-get} have extended the pipeline.
Regarding scientific claims, \citet{wadden-etal-2022-multivers} showed that full-document context
with weak supervision substantially outperforms sentence-level approaches.
Fact-checking has also been applied to multi-hop settings~\cite{feverous,jiang-etal-2020-hover},
news headlines~\cite{popat-etal-2018-declare}, COVID-19 claims~\cite{saakyan-etal-2021-covid},
and LLM-generated text~\cite{min-etal-2023-factscore}.
The explainability of AFC systems has also gained attention, including for health claims~\cite{kotonya-toni-2020-explainable}. Annual AFC shared tasks continue to push that progress~\cite{fever2025, climatecheck2025}.

Recent progress in AFC has introduced agentic systems, like FIRE~\cite{xie-etal-2025-fire} and DEFAME~\cite{braun2025defame}. FIRE jointly performs evidence retrieval and claim verification, dynamically issuing request for additional search based on verifier's confidence. DEFAME extends this idea by adding visual evidence. Such systems shifted from retrieve-then-verify paradigm and allow verifier to control evidence retrieval.

\paragraph{Fact-checking benchmarks.}
LIAR~\cite{wang-2017-liar} and MultiFC~\cite{augenstein-etal-2019-multifc} collect
real-world claims from political fact-checking organisations, but do not provide structured
evidence corpora.
\citet{scifact} introduced \scifact{} for verifying scientific claims against biomedical
abstracts from S2ORC~\cite{s20rc}, requiring domain knowledge beyond lexical matching.
\citet{climatefever} developed \climatefever{} by linking climate claims to Wikipedia passages, focusing on multi-sentence reasoning.
\citet{averitec} proposed \averitec{}, in which evidence is retrieved from the live web at
claim time, posing a realistic but difficult open-domain retrieval problem.
\citet{climatecheck} introduced \climatecheck{}, linking social-media climate posts to scientific articles and combining the challenges of informal language and a 394K-document corpus.

\paragraph{Evidence retrieval.}
Early retrievers used sparse models, like TF-IDF and BM25~\cite{Robertson2009bm25}.
Dense passage retrieval (DPR)~\cite{karpukhin-etal-2020-dense}, Sentence-BERT~\cite{reimers-gurevych-2019-sentence},
ColBERT~\cite{khattab2020colbert}) showed improvement over sparse methods on in-domain benchmarks.
Retrieval-augmented generation (RAG)~\cite{lewis2020retrieval} combines generative models with evidence grounding, further pushing retrieval performance. However, specialised training is still required for domain-specific AFC corpora, which are rarely available.

\paragraph{Benchmarking studies.}
\citet{calamai2025benchmarking} performed a reproducibility study across 29 climate-related
NLP datasets, finding that classical baselines rarely fall far behind fine-tuned models,
that most datasets contain annotation issues,
and that performance differences are often within statistical confidence intervals.
\citet{thakur2021beir} found a parallel result for information retrieval:
BM25 outperforms dense models trained on MS MARCO across the majority of heterogeneous corpora.
To the best of our knowledge, prior studies did not analyse multi-stage AFC pipelines across diverse datasets.

\section{Tasks, Datasets, and Models}
\label{sec:benchmark}

\begin{table*}[t]
  \centering
  \small
  \renewcommand{\arraystretch}{1.15}
  \begin{tabular}{lllrrl}
    \toprule
    \textbf{Dataset} & \textbf{Domain} & \textbf{Claim type} & \textbf{Claims}
      & \textbf{Evidence corpus} & \textbf{Labels} \\
    \midrule
    \rowcolor{green!10}\averitec{}     & Open-web        & Real-world          & 5{,}783
      & 32{,}818 web docs        & Sup / Ref / NEI / Conflicting \\
    \rowcolor{blue!10}\scifact{}      & Life sciences   & Expert-written      & 1{,}409
      & 5{,}183 S2ORC abstracts  & Supports / Refutes / NEI \\
    \rowcolor{orange!12}\climatecheck{} & Climate/social  & Social-media        & 3{,}199
      & 394{,}269 sci.\ abstracts & Supports / Refutes / NEI \\
    \rowcolor{violet!8}\climatefever{} & Climate science & Real-world          & 7{,}675
      & 5{,}240 Wikipedia pages  & Sup / Ref / NEI / Conflicting \\
    \bottomrule
  \end{tabular}
  \caption{Overview of the datasets used in the study.}
  \label{tab:datasets}
\end{table*}

\subsection{Task}
\label{sec:task}

\begin{figure}[t]
\centering
\resizebox{\columnwidth}{!}{%
\begin{tikzpicture}[
  s1box/.style={draw=teal!70!black, rounded corners=4pt, minimum width=2.1cm,
                minimum height=0.68cm, align=center, font=\small,
                fill=teal!18, thick},
  s2box/.style={draw=orange!70!black, rounded corners=4pt, minimum width=2.1cm,
                minimum height=0.68cm, align=center, font=\small,
                fill=orange!22, thick},
  cbox/.style={draw=blue!50!black, rounded corners=4pt, minimum width=2.1cm,
               minimum height=0.68cm, align=center, font=\small,
               fill=blue!14, thick},
  sbox/.style={draw=gray!60, rounded corners=3pt, minimum width=2.3cm,
               minimum height=0.58cm, align=center, font=\small,
               fill=gray!10, densely dashed},
  arr1/.style={-{Stealth[length=5pt,width=4pt]}, thick, color=teal!70!black},
  arr2/.style={-{Stealth[length=5pt,width=4pt]}, thick, color=orange!70!black},
  darr/.style={-{Stealth[length=5pt,width=4pt]}, thick, dashed, gray!60},
  lbl1/.style={font=\scriptsize\itshape, text=teal!70!black},
  lbl2/.style={font=\scriptsize\itshape, text=orange!80!black}
]
  \node[cbox]  (claim)  {Claim $c$};
  \node[s1box, right=0.95cm of claim] (ret)   {Retrieval};
  \node[s1box, right=0.95cm of ret]   (topk)  {Evidence $\hat{E}_K$};
  \node[s2box, right=0.95cm of topk]  (ver)   {Veracity};
  \node[s2box, right=0.9cm  of ver]   (lbl)   {Label $y$};
  \node[sbox,  below=0.62cm of ret]   (corpus){Corpus $\mathcal{D}$};
  \node[lbl1, above=0.08cm of ret]  {Stage 1};
  \node[lbl2, above=0.08cm of ver]  {Stage 2};
  \draw[arr1] (claim)  -- (ret);
  \draw[arr1] (corpus) -- (ret);
  \draw[arr1] (ret)    -- (topk);
  \draw[arr2] (topk)   -- (ver);
  \draw[arr2] (ver)    -- (lbl);
  \draw[darr] (claim.south) -- ++(0,-0.30) -| (ver.south);
\end{tikzpicture}%
}%
\caption{Classical fact-checking pipeline. \textcolor{teal!70!black}{Stage~1} retrieves
  evidence $\hat{E}_K$ from corpus $\mathcal{D}$; \textcolor{orange!80!black}{Stage~2}
  predicts veracity label $y$.}
\label{fig:pipeline}
\end{figure}
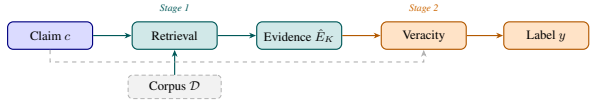

A classical simplified pipeline has retriever and verifier modules (illustrated in Figure~\ref{fig:pipeline}).
Given a claim $c$ and an evidence corpus $\mathcal{D}$, a system should:
(1)~\textbf{retrieve} the $K$ most relevant documents $\hat{E} \subseteq \mathcal{D}$;
and (2)~\textbf{predict} veracity $y \in \{\textit{Supports},\,\textit{Refutes},\,\textit{NEI}\}$
given $c$ and $\hat{E}$.
We evaluate the two stages separately in Section~\ref{sec:evaluation} and jointly in Table~\ref{tab:oracle} (Section~\ref{sec:res}).

\subsection{Datasets}
\label{sec:datasets}

Table~\ref{tab:datasets} and Figure~\ref{fig:taxonomy} summarise four datasets used in this study.
The datasets cover commonsense and political claims from the open web (\averitec{}),
climate claims from journalism and Wikipedia (\climatefever{}),
expert-written scientific claims (\scifact{}),
and user-generated social-media climate claims (\climatecheck{}).
This organisation by evidence domain and claim origin (Figure~\ref{fig:taxonomy})
reveals two key structural axes that drive the retrieval results in Section~\ref{sec:results}:
datasets in the \emph{scientific evidence} row require semantic matching beyond lexical overlap, while datasets in the \emph{real-world/social} column contain informal language that widens the vocabulary gap between claim and evidence.

\begin{figure}[t]
\centering
\scriptsize
\setlength{\tabcolsep}{2pt}%
\renewcommand{\arraystretch}{1.3}%
\begin{tabular}{>{\bfseries\centering\arraybackslash}p{1.4cm}%
  |>{\centering\arraybackslash}p{2.85cm}%
  |>{\centering\arraybackslash}p{2.85cm}|}
\cline{2-3}
\multicolumn{1}{c|}{} & \textbf{Expert / curated} & \textbf{Real-world / social} \\
\hline
Scientific
  & \cellcolor{blue!16}\textbf{\scifact{}}\newline
    1.4K claims\newline 5K abstracts\newline biomedical domain
  & \cellcolor{orange!20}\textbf{\climatecheck{}}\newline
    3.2K claims\newline 394K abstracts\newline social-media lang. \\[2pt]
\hline
Web / Wiki
  & \cellcolor{green!16}\textbf{\averitec{}}\newline
    5.8K claims\newline 33K web docs\newline open-web evidence
  & \cellcolor{violet!12}\textbf{\climatefever{}}\newline
    7.7K claims\newline 5K passages\newline Wikipedia sents. \\[2pt]
\hline
\end{tabular}
\caption{Dataset taxonomy by evidence type (rows) and claim origin (columns).
  Each dataset is color-coded consistently throughout the paper.}
\label{fig:taxonomy}
\end{figure}

We apply an 80/10/10 train/development/test split and retain original splits when already provided in these proportions.
Dataset descriptions, data quality checks (gibberish detection, duplicate removal, text cleaning), split sizes, and label distributions available in Appendix~\ref{app:dataset-desc}.

\subsection{Models}
\label{sec:models}

We evaluate a range of models on each task: random baselines, sparse retrieval methods,
fine-tuned transformers, zero-shot LLMs, and state-of-the-art shared-task systems.

\paragraph{Evidence retrieval.}
\textbf{\sys{Random}} selects $K$ documents uniformly at random.
\textbf{\sys{TF-IDF}}~\cite{manning2008introduction} ranks documents by cosine similarity
on TF-IDF vectors.
\textbf{\sys{BM25}}~\cite{Robertson2009bm25} extends TF-IDF with document-length
normalisation using the Okapi BM25 scoring function (see Appendix~\ref{app:metrics}
for the full formula).
\textbf{\sys{AIC CTU}}~\cite{aic-ctu} is the \averitec{} 2025 shared task winner, which
uses long-context RAG~\cite{lewis2020retrieval} over retrieved passages.
\textbf{\sys{Sanctuary}}~\cite{sanctuary} is the runner-up from the same shared task, combining dense retrieval with neural reranking.
Both systems were developed and evaluated on FEVER-style encyclopaedic claims, and are best under \averitec{}~2025 shared-task constraints~\cite{akhtar-etal-2025-2nd} (open-weights, $\leq23$ GB GPU, $\leq1$ minute per claim, fixed evidence store, no closed-weight LLMs); they are therefore strongest \emph{under those constraints} rather than in absolute terms, and stronger unconstrained systems exist.

\paragraph{Claim veracity prediction.}
We separated baselines for claim-verification task into two classes: \textit{claim-/hypothesis-only} baselines and 2) \textit{evidence-conditioned}. Following hypothesis-only analyses in NLI~\cite{poliak-etal-2018-hypothesis} and claim-only analyses in fact verification~\cite{schuster-etal-2019-towards}, claim-only condition is introduced to quantify whether labels can be predicted from the claim without access to evidence. This condition is intended as a diagnostic probe for dataset artifacts or annotation shortcuts.

Logistic regression over TF-IDF/BM25 features and fine-tuned transformer models are claim-only baselines. Zero-shot LLMs are tested in claim-only and evidence-conditioned settings over \sys{BM25} retrieval results.

\textbf{\sys{Random}} predicts a label drawn uniformly at random.
\textbf{\sys{TF-IDF + LogReg}} and \textbf{\sys{BM25 + LogReg}} use TF-IDF or BM25 vectors based on claim as input to a logistic regression classifier~\cite{pedregosa2011scikit}.
\textbf{\sys{Longformer}}~\cite{Beltagy2020Longformer} and
\textbf{\sys{DistilRoBERTa}}~\cite{Sanh2019DistilBERTAD} are two transformer~\cite{vaswani2017attention} models fine-tuned for sequence classification. 
LLMs include \textbf{\sys{Llama~3.1-8B}} and \textbf{\sys{Llama~3.1-70B}}~\cite{llama3modelcard}.
Task-specific best performing systems are \textbf{\sys{Sanctuary}} and \textbf{\sys{AIC CTU}}, which
use their own retrieval components.

\subsection{Evaluation}
\label{sec:evaluation}

\paragraph{Evidence retrieval.}
We report Recall@$K$ and F1@$K$ for $K \in \{5, 10, 20\}$, computed over the set of
gold-annotated relevant documents.
Recall@$K$ measures the fraction of gold-relevant documents recovered in the top-$K$
results; it plateaus once $K$ exceeds the gold set size, which affects \averitec{}
(see Section~\ref{sec:res}).
A retrieved document is relevant if annotated as \emph{Supports} or \emph{Refutes};
formal metric definitions are in Appendix~\ref{app:metrics}.
For \averitec{}, we follow the official shared-task protocol:
evidence is evaluated via the Hungarian METEOR score~\cite{hungarian,meteor}
(see Appendix~\ref{app:metrics} for the full definition), with a threshold of 0.25
to match the QA-pair format of that dataset.

\paragraph{Claim veracity prediction.}
We report accuracy and macro-averaged F1.
Macro F1 averages class-level F1 scores equally across $C$ classes to address dominant veracity label classes. A high level of imbalance may cause accuracy and macro F1 to diverge by up to 20 points for the same model.
Evaluation is based on annotated claim-evidence pairs, and unjudged passages are ignored.

\paragraph{Statistical significance.}
We compute 95\% confidence intervals (CI) of macro-F1 scores via bootstrapping~\cite{efron1994introduction}.
A difference is considered significant if the CIs are disjoint.

\section{Experimental Results and Analysis}
\label{sec:results}

\subsection{Experimental Setup}
\label{sec:experiments}

All experiments use the same pre-processed splits described in Section~\ref{sec:datasets}.
We use three fixed random seeds and report means $\pm$ standard deviation;
single-run results are reported for LLMs (temperature 0.1).
Full implementation details, hyperparameters, prompt templates, and hardware information
are in Appendix~\ref{app:supplementary} and~\ref{app:prompts}.

\subsection{Results}
\label{sec:res}

\begin{table*}[t]
\centering
\setlength{\tabcolsep}{4.5pt}
\renewcommand{\arraystretch}{1.18}
\begin{tabular}{ll cccccc}
\toprule
\multirow{2}{*}{\textbf{Dataset}}
  & \multirow{2}{*}{\textbf{Method}}
  & \multicolumn{3}{c}{\textbf{Recall}}
  & \multicolumn{3}{c}{\textbf{F1}}\\
\cmidrule(lr){3-5}\cmidrule(lr){6-8}
 & & \textbf{@5} & \textbf{@10} & \textbf{@20}
   & \textbf{@5} & \textbf{@10} & \textbf{@20} \\
\midrule

\multirow{5}{*}{{\averitec{}}}
  & \sys{Random}
    & 0.0054 & 0.0054 & 0.0054
    & 0.0026 & 0.0016 & 0.0009 \\
  & \sys{TF-IDF}
    & \textbf{0.1258} & \textbf{0.1258} & 0.1258
    & \textbf{0.0652} & \textbf{0.0384} & 0.0245 \\
  & \sys{BM25}
    & {0.1170} & {0.1170} & {0.1170}
    & {0.0577} & {0.0341} & {0.0218} \\
  & \sys{Sanctuary}
    & \textcolor{gray}{0.0769} & \textcolor{gray}{0.1195} & \textbf{0.1694}
    & \textcolor{gray}{0.0373} & \textcolor{gray}{0.0361} & \textbf{0.0300} \\
  & \sys{AIC CTU}
    & {0.0749} & {0.0785} & {0.0785}
    & {0.0372} & {0.0227} & {0.0124} \\
\midrule

\multirow{5}{*}{{\scifact{}}}
  & \sys{Random}
    & 0.0000 & 0.0067 & 0.0067
    & 0.0000 & 0.0012 & 0.0006 \\
  & \sys{TF-IDF}
    & 0.2174 & 0.4903 & {0.8053}
    & 0.0774 & 0.0967 & 0.0848 \\
  & \sys{BM25}
    & {0.2128} & {0.4642} & {0.7458}
    & {0.0750} & {0.0901} & {0.0767} \\
  & \sys{Sanctuary}
    & \underline{0.6501} & \underline{0.6759} & \textcolor{gray}{0.6804}
    & \underline{0.4156} & \underline{0.4025} & \underline{0.3984} \\
  & \sys{AIC CTU}
    & \textbf{0.7306} & \textbf{0.7426} & \underline{0.7453}
    & \textbf{0.4678} & \textbf{0.4596} & \textbf{0.4623} \\
\midrule

\multirow{5}{*}{{\climatecheck{}}}
  & \sys{Random}
    & 0.0000 & 0.0000 & 0.0000
    & 0.0000 & 0.0000 & 0.0000 \\
  & \sys{TF-IDF}
    & 0.0795 & {0.1253} & {0.2074}
    & 0.0346 & 0.0306 & 0.0269 \\
  & \sys{BM25}
    & {0.0646} & {0.1195} & {0.1802}
    & {0.0286} & {0.0303} & {0.0240} \\
  & \sys{Sanctuary}
    & \underline{0.1161} & \textbf{0.1766} & \textbf{0.2596}
    & \underline{0.0374} & \underline{0.0361} & \underline{0.0318} \\
  & \sys{AIC CTU}
    & \textbf{0.1022} & \textcolor{gray}{0.1170} & \textcolor{gray}{0.1229}
    & \textbf{0.0559} & \textbf{0.0545} & \textbf{0.0545}  \\
\midrule

\multirow{5}{*}{{\climatefever{}}}
  & \sys{Random}
    & 0.0013 & 0.0026 & 0.0091
    & 0.0013 & 0.0017 & 0.0036 \\
  & \sys{TF-IDF}
    & 0.1377 & 0.2260 & {0.3429}
    & 0.1454 & {0.1789} & {0.1949} \\
  & \sys{BM25}
    & {0.1000} & {0.1805} & {0.2844}
    & {0.1057} & {0.1434} & {0.1634}\\
  & \sys{Sanctuary}
    & \underline{0.1925} & \underline{0.2618} & \textcolor{gray}{0.3034}
    & \underline{0.1626} & \textcolor{gray}{0.1677} & \textcolor{gray}{0.1597} \\
  & \sys{AIC CTU}
    & \textbf{0.2425} & \textbf{0.3011} & \textbf{0.3450}
    & \textbf{0.2434} & \textbf{0.2207} & \textbf{0.2202} \\

\bottomrule
\end{tabular}
\caption{
  Evidence retrieval results (Recall@$K$ and F1@$K$).
  \textbf{Bold}: best per column.
  \underline{Underline}: second best.
  \textcolor{gray}{Gray}: not significantly better than \sys{TF-IDF}.
  For \averitec{}, the official Hungarian METEOR ($\geq 0.25$) is used instead of exact-match relevance.
  $\dagger$~For sparse methods on \averitec{}, Recall@$K$ is identical across $K\in\{5,10,20\}$ because
  the median gold set contains a single relevant document; once retrieved in the top-5, increasing $K$ yields no further coverage.
}
\label{tab:retrieval_results}
\end{table*}

\begin{table*}[t]
\centering
\begin{tabular}{ll ccc}
\toprule
\textbf{System} & \textbf{Dataset} & \textbf{Q only}  & \textbf{Q + A} & \textbf{AVeriTeC Score} \\
\midrule
\multirow{4}{*}{{\sys{Sanctuary} }}
  & \averitec{} & 0.4461 & 0.2152 & 0.2200 \\
  & \scifact{} & 0.6603 & 0.4343 & 0.4900 \\
  & \climatecheck{} & 0.4614 & 0.3142 & 0.4601 \\
  & \climatefever{} & 0.4696 & 0.3518 & 0.3247 \\
\midrule
\multirow{4}{*}{{\sys{AIC CTU} }}
  & \averitec{} & 0.4610 & 	0.3264 & 0.5360 \\
  & \scifact{} & 0.5071 & 0.2509 & 0.3900 \\
  & \climatecheck{} & 0.3404 & 0.0586 & 0.0000 \\
  & \climatefever{} & 0.3443 & 0.2081 & 0.1494 \\
\bottomrule
\end{tabular}
\caption{\sys{Sanctuary} and \sys{AIC CTU} performance under the official \averitec{} evaluation protocol
  (Hungarian METEOR $\geq 0.25$; see Appendix~\ref{app:metrics}).
  \textbf{Purpose:} applying this metric cross-dataset reveals how effectively systems
  retrieve evidence whose text semantically matches the annotated question--answer pairs,
  isolating evidence adequacy from label prediction accuracy.
  \textbf{Q~only:} METEOR score computed on retrieved question text alone (measures
  query coverage); \textbf{Q~+~A:} METEOR computed on question plus answer text
  (measures full evidence adequacy).
  Scores on non-\averitec{} datasets are lower because those datasets lack QA-pair
  annotations, but the Q~only vs.\ Q~+~A gap still quantifies how much answer content
  contributes to evidence quality for each system.}
\label{tab:averitec_results}
\end{table*}

\begin{table*}[t]
\centering
\setlength{\tabcolsep}{2.5pt}
\renewcommand{\arraystretch}{1.15}
\resizebox{\textwidth}{!}{%
\begin{tabular}{lc rrrr rrrr r}
\toprule
  & & \multicolumn{2}{c}{\textbf{\averitec{}}}
    & \multicolumn{2}{c}{\textbf{\scifact{}}}
    & \multicolumn{2}{c}{\textbf{\climatecheck{}}}
    & \multicolumn{2}{c}{\textbf{\climatefever{}}}
    & \textbf{Avg.} \\
\cmidrule(lr){3-4}\cmidrule(lr){5-6}\cmidrule(lr){7-8}\cmidrule(lr){9-10}
\textbf{Method} & \textbf{Ev.}
  & \textbf{Acc} & \textbf{F1}
  & \textbf{Acc} & \textbf{F1}
  & \textbf{Acc} & \textbf{F1}
  & \textbf{Acc} & \textbf{F1} & \textbf{F1} \\
\midrule
\multicolumn{10}{l}{\textit{Baselines}} \\[1pt]
\sys{Random}            & \pmark
  & \textcolor{gray}{\ms{27.90}{2.13}} & \textcolor{gray}{\ms{23.32}{2.02}}
  & \textcolor{gray}{\ms{33.37}{2.17}} & \textcolor{gray}{\ms{33.17}{2.24}}
  & \textcolor{gray}{\ms{34.94}{2.05}} & \textcolor{gray}{\ms{32.78}{1.72}}
  & \textcolor{gray}{\ms{24.19}{3.61}} & \textcolor{gray}{\ms{21.84}{2.7}}
  & 0.2778 \\
\sys{TF-IDF + LogReg}   & \pmark
  & 55.20 & 37.06
  & 42.67 & 39.63
  & 60.17 & 59.45
  & 39.61 & 36.84
  & 43.25 \\
\sys{BM25 + LogReg}     & \pmark
  & \textcolor{gray}{54.80} & \textcolor{gray}{32.15}
  & \textcolor{gray}{45.67} & \textcolor{gray}{42.34}
  & \textcolor{gray}{57.52} & \textcolor{gray}{56.90}
  & {44.45} & \underline{38.64}
  & 42.51 \\
\midrule
\multicolumn{10}{l}{\textit{Fine-tuned transformers}} \\[1pt]
\sys{DistilRoBERTa}     & \pmark
  & \textcolor{gray}{\ms{56.20}{2.47}} & \textcolor{gray}{\ms{37.39}{2.60}}
  & \textcolor{gray}{\ms{45.56}{3.50}} & \textcolor{gray}{\ms{42.45}{2.16}}
  & \underline{\ms{61.65}{0.84}} & \underline{\ms{60.48}{0.80}}
  & \ms{43.01}{1.52} & \textcolor{gray}{\ms{32.34}{5.67}}
  & 43.17 \\
\sys{Longformer}        & \pmark
  & \textcolor{gray}{\ms{57.53}{3.09}} & \textcolor{gray}{\ms{37.77}{2.77}}
  & \textcolor{gray}{\ms{46.11}{0.57}} & \textcolor{gray}{\ms{43.81}{0.36}}
  & \textbf{\ms{61.82}{0.59}} & \textbf{\ms{60.67}{0.74}}
  & \ms{44.09}{6.63} & \textcolor{gray}{\ms{29.25}{4.83}}
  & 42.88 \\
\midrule
\multicolumn{10}{l}{\textit{Zero-shot LLMs}} \\[1pt]
\sys{Llama 8B}      & \pmark
  & \textcolor{gray}{31.00} & \textcolor{gray}{28.63}
  & \textcolor{gray}{46.00} & \textcolor{gray}{38.65}
  & \textcolor{gray}{52.56} & \textcolor{gray}{46.53}
  & \textcolor{gray}{38.96} & \textcolor{gray}{30.10}
  & 36.00 \\
\sys{Llama 70B}     & \pmark
  & {64.00} & {43.36}
  & \textcolor{gray}{47.00} & \textcolor{gray}{39.30}
  & \textcolor{gray}{53.88} & \textcolor{gray}{44.84}
  & \textcolor{gray}{24.03} & \textcolor{gray}{17.46}
  & 36.24 \\
\sys{BM25 + Llama 8B}      & \cmark
  & \textcolor{gray}{52.60} & \textcolor{gray}{39.50}
  & \textcolor{gray}{46.33} & \textcolor{gray}{41.09}
  & \textcolor{gray}{28.17} & \textcolor{gray}{28.07}
  & {42.86} & {36.28}
  & 36.24 \\
\sys{BM25 + Llama 70B}      & \cmark
  & \underline{70.80} & {43.18}
  & \textcolor{gray}{47.33} & \textcolor{gray}{44.86}
  & \textcolor{gray}{34.51} & \textcolor{gray}{31.99}
  & {44.16} & {34.68}
  & 38.68 \\
\midrule
\multicolumn{10}{l}{\textit{SOTA systems}} \\[1pt]
\sys{Sanctuary}         & \cmark
  & \textbf{\ms{70.93}{0.19}} & \textbf{\ms{48.18}{0.39}}
  & \textbf{\ms{70.22}{0.42}} & \textbf{\ms{70.00}{0.43}}
  & \textcolor{gray}{\ms{37.02}{0.17}} & \textcolor{gray}{\ms{31.45}{0.24}}
  & \underline{\ms{44.59}{0.81}} & \textbf{\ms{38.92}{0.68}}
  & 47.14 \\
\sys{AIC CTU}           & \cmark
  & {\ms{60.60}{0.12}} & \textcolor{gray}{\ms{37.14}{0.37}}
  & \underline{\ms{56.56}{0.16}} & \underline{\ms{48.22}{0.47}}
  & \textcolor{gray}{\ms{39.72}{0.55}} & \textcolor{gray}{\ms{28.73}{0.74}}
  & \textbf{\ms{45.24}{0.81}} & {\ms{33.60}{1.23}}
  & 36.92 \\
\bottomrule
\end{tabular}}
\caption{
  Veracity prediction results (mean $\pm$ std over 3 seeds where applicable;
  single run for LLMs).
  \textbf{Ev.:} \cmark~=~claim + retrieved evidence;
  \pmark~=~claim-only (no retrieved evidence).
  \textbf{Bold}: best per column.
  \underline{Underline}: second best.
  \textcolor{gray}{Gray}: not significantly better than \sys{TF-IDF + LogReg}
  (bootstrap 95\% CI, $p{<}0.05$).
  Full per-class breakdown in Table~\ref{tab:per_class}.
}
\label{tab:veracity_summary}
\end{table*}

\paragraph{Evidence retrieval.}
An interesting pattern in Table~\ref{tab:retrieval_results} is the reversal of rankings between datasets.
On \scifact{}, \sys{AIC CTU} and \sys{Sanctuary} dramatically outperform sparse methods:
\sys{AIC CTU} achieves R@5 = 0.731 versus \sys{TF-IDF}'s 0.217.
The scientific vocabulary of \scifact{} abstracts rewards dense semantic
retrieval.
On \averitec{}, the rankings change: \sys{TF-IDF} leads at R@5 (0.126) while \sys{Sanctuary} reaches only 0.077.
Open-web evidence is better matched by lexical overlap than by dense representations trained on encyclopaedic claims, supporting findings in heterogeneous IR benchmarks~\cite{thakur2021beir}.
On \climatecheck{} and \climatefever{}, retrieval results are low for all methods, demonstrating the challenges with large or informal corpora.
Additionally, \sys{TF-IDF} consistently outperforms \sys{BM25} across all datasets. This is likely caused by BM25's document-length normalisation penalty, putting long Wikipedia passages and scientific abstracts at a disadvantage.

Table~\ref{tab:averitec_results} shows system performance under the official \averitec{} metrics, where \sys{AIC CTU} leads with a score of 0.536 on \averitec{} dataset. However, \sys{Sanctuary} outperforms on the rest of the datasets on all metrics. For instance, on \climatecheck{} \sys{AIC CTU} shows near-zero \averitec{} score, compared to 0.46 scored by \sys{Sanctuary}. This is likely caused by architectural design choices, where \sys{AIC CTU} creates FAISS vectors for all evidence documents, which works well for \averitec{} that has 32,818 web documents, but does not scale for \climatecheck{} with its 394K abstracts.

\paragraph{Claim veracity prediction.}
Table~\ref{tab:veracity_summary} reports accuracy and macro-F1 for all models. 
On \climatecheck{}, fine-tuned \sys{Longformer} achieves the highest accuracy (0.618), followed by \sys{DistilRoBERTa} (0.617). 
Surprisingly, evidence can hurt under domain shift, adding BM25 retrieved evidence decreases accuracy for LLMs. While \sys{AIC CTU} degradation on \climatecheck{} could be attributed to the challenges of scaling retreval configuration to a large corpus.
This confirms that low-quality evidence retrieval misleads the veracity model when the domain gap between informal claims and scientific evidence is large, hence performances of top systems are also below \sys{TF-IDF + LogReg} baseline. The high claim-only performance suggests that \climatecheck{} contains stronger exploitable correlations between claim text and labels.

On \climatefever{}, all methods struggle, with peak accuracy of 0.45 (\sys{AIC CTU}), followed by \sys{Sanctuary} (0.446) and \sys{Longformer} (0.441). A small difference between simple and complex models demonstrates that evidence quality is the key factor. All systems show a large gap between accuracy and macro F1 suggesting heavy class imbalance. \emph{Supports} label dominates and \emph{Conflicting Evidence} is sparse.

\scifact{} shows a clear winning model: \sys{Sanctuary} leads with accuracy 0.702 and macro F1 0.700, followed by \sys{AIC CTU} (accuracy 0.566, macro F1 0.482), while all fine-tuned models cluster below 0.47 accuracy and are not significantly better than \sys{TF-IDF + LogReg}. \sys{Sanctuary}'s strong performance is likely due to architectural design. Evidence documents are chunked at sentence-level, then semantically grouped to form a coherent evidence unit, which aligns well with \scifact{}'s rationale-sentence annotations.
The accuracy-macro~F1 gap is the smallest of the four datasets, reflecting well-balanced, expert-controlled annotations.

On \averitec{}, \sys{Sanctuary} leads (accuracy 0.709, macro F1 0.482), while \sys{BM25 + Llama 70B} comes second (accuracy 0.708).
Fine-tuned transformers reach 0.562-0.575 accuracy, a moderate but clear gap to strong evidence-conditioned baselines. Large gaps between accuracy and macro F1 confirm skewed class distribution, with \emph{NEI} and \emph{Conflicting Evidence} having minimal examples.

Claim-only performance provides evidence of dataset-specific shortcut signals. Relative to random prediction, claim-only macro-F1 improves by 13.7 points on \averitec{}, 6.5 on \scifact{}, 26.7 on \climatecheck{}, and 15.0 on \climatefever{}. The particularly large gain on \climatecheck{} indicates that substantial label-predictive information is available from claim text alone. 
The datasets on which claim-only baselines come closest to, or exceed, evidence-conditioned systems are also those with the weakest evidence annotations: \citet{calamai2025benchmarking} report Cohen's $\kappa = 0.334$ for \climatefever{} evidence annotations, and our own two-annotator re-labelling of sampled errors reaches only $\kappa = 0.55$ (Appendix~\ref{app:error-analysis}). Part of the apparent baseline advantage therefore reflects the benchmark limitations we set out to measure, and we treat gaps below ${\sim}5$ points as within the annotation-noise floor.

The dominant cross-dataset pattern is rank instability. Fine-tuned transformers are the most reliable non-SOTA systems, whereas zero-shot LLMs exhibit high variance, competitive on \averitec{} (\sys{Llama~70B} accuracy 0.640) but near-random on \climatefever{} (\sys{Llama~70B} accuracy 0.240).
\sys{TF-IDF + LogReg} shows strong performance over zero-shot LLMs and \averitec{} 2025 top-performing systems on \climatecheck{} , the largest and most informal corpus, suggesting that term-frequency features are sufficient when neural systems are out-of-domain.

Additionally we examined the climate-related subset of \averitec{}. Only six claims (1.2\%) are climate-related, which is too small to support a meaningful topic-controlled comparison. Moreover, \sys{Sanctuary}  does not show a corresponding performance degradation on these claims (83.3 vs 70.9). We therefore do not interpret this subset as evidence isolating topic effects.

\subsection{Error Analysis}
\label{sec:analysis}

Following \citet{calamai2025benchmarking}, we sampled errors across all four datasets and
classified them as genuine model errors, annotation mistakes, or debatable errors.
The dominant failure modes are vocabulary mismatch on \climatecheck{} (informal claim language
vs.\ formal abstracts), label ambiguity at the Refutes/NEI boundary across all datasets, and domain mismatch in storng evidence-conditioned baslines fine-tuned on encyclopaedic claims.
Full per-failure-mode analysis is in Appendix~\ref{app:error-analysis}.

To quantify how much veracity performance is limited by retrieval quality rather than the veracity model itself, we re-evaluate all models using gold-annotated evidence instead of \sys{TF-IDF}-retrieved documents (Table~\ref{tab:oracle}). Across LLM models, replacing retrieved evidence with gold evidence improves accuracy by 14-22 percentage points.
The largest gains appear for \sys{Llama~70B} on \climatefever{} ($+$22~pp) and \sys{Llama~8B} on \climatefever{} ($+$20~pp), confirming that multi-passage retrieval failures are the main bottleneck on that dataset.
The smallest oracle gains are on \climatecheck{} ($+$14-15~pp), where the large corpus limits coverage even when gold relevance is assumed.

Prior works have already questioned whether fact-checking models genuinely rely on evidence for their predictions~\cite{hansen-etal-2021-automatic, schuster-etal-2019-towards}. Hence, Table~\ref{tab:identical-setting} demonstrates such an ablation study with identical setting on two LLMs. On \averitec{} the evidence helps, while on \climatecheck{} it hurts.
Since the model and prompt are identical in both settings, the classical baseline's advantage is a real failure on out-of-domain conditions, rather than superiority of claim-only models.
This conclusion is not specific to open-weight verifiers. Appendix~\ref{app:frontier} repeats the ablation with a frontier closed-weight model under three evidence conditions, where retrieved evidence again scores below claim-only input, and Appendix~\ref{app:agentic} tests an iterative \sys{FIRE}-style~\cite{xie-etal-2025-fire} agent that improves over one-shot retrieval on \averitec{} but remains far below oracle retrieval on both datasets.

In addition, to quantify lexical compatibility between claims and evidence, we compute claim–evidence word-overlap using Jaccard similarity (Table~\ref{tab:jaccard}). Mean overlap is 0.122 on \climatefever{},  0.087 on \scifact{}, 0.093 on \averitec{}, and only 0.047 on \climatecheck{}. Low lexical overlap on \climatecheck{} is consistent with the larger vocabulary mismatch between informal claims and scientific abstracts, providing a measurable correlate of the retrieval difficulty observed in this dataset.

\begin{table}[t]
\centering
\small
\setlength{\tabcolsep}{5pt}
\begin{tabular}{lcccc}
\toprule
\textbf{Model} & \textbf{AVT} & \textbf{SCI} & \textbf{CCK} & \textbf{CFV} \\
\midrule
\sys{Llama~8B}     & $+$16 & $+$19 & $+$14 & $+$20 \\
\sys{Llama~70B}    & $+$14 & $+$18 & $+$15 & $+$22 \\
\bottomrule
\end{tabular}
\caption{Accuracy improvement (percentage points) when gold evidence replaces retrieved evidence.
AVT~=~\averitec{}; SCI~=~\scifact{}; CCK~=~\climatecheck{}; CFV~=~\climatefever{}.
}
\label{tab:oracle}
\end{table}

\begin{table}[t]
\centering
\small
\setlength{\tabcolsep}{5pt}
\begin{tabular}{lcccc}
\toprule
\textbf{Model} & \textbf{AVT} & \textbf{SCI} & \textbf{CCK} & \textbf{CFV} \\
\midrule
\sys{Llama~8B}     & $+$21.6 & $+$0.3 & $-$24.4 & $+$3.9 \\
\sys{Llama~70B}    & $+$6.8 & $+$0.3 & $-$19.4 & $+$20.1 \\
\bottomrule
\end{tabular}
\caption{A matched, identical-setting ablation (same model and prompt, ± retrieved evidence). Accuracy change when the same model is given retrieved evidence instead of claim-only.
}
\label{tab:identical-setting}
\end{table}

\begin{table}[t]
\centering
\small
\setlength{\tabcolsep}{5pt}
\begin{tabular}{lc}
\toprule
\textbf{Dataset} & \textbf{Mean Jaccard} \\
\midrule
\averitec{}     & 0.093 \\
\scifact{}      & 0.087 \\
\climatecheck{} & 0.047 \\
\climatefever{} & 0.122 \\
\bottomrule
\end{tabular}
\caption{Jaccard similarity analysis. Per-dataset mean word overlap between claims and gold evidence.}
\label{tab:jaccard}
\end{table}

\subsection{Discussion}
\label{sec:discussion}

\begin{figure}[t]
  \includegraphics[width=\columnwidth]{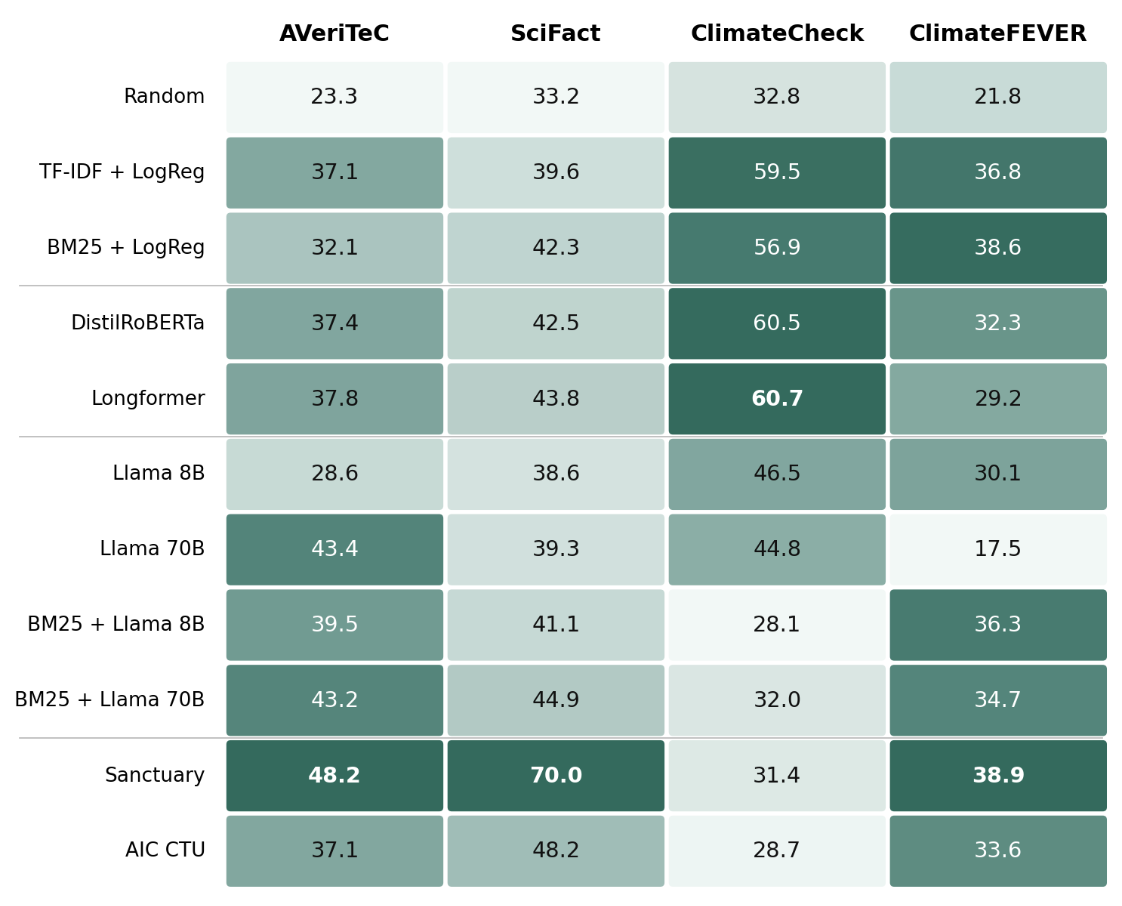}
  \caption{Macro-F1 heatmap for every method-dataset pair on the claim veracity task.}
  \label{fig:heatmap-main}
\end{figure}

\paragraph{System rankings are dataset-specific.}
No single system dominates across all four benchmarks. 
Macro-F1 rankings in Figure~\ref{fig:heatmap-main} make this vivid: \sys{Sanctuary} shows the darkest cells on \scifact{} (0.700) and \averitec{} (0.482) but the lightest on \climatecheck{} (0.315).
Similarly, the \averitec{}~2025 winner \sys{AIC CTU} and runner-up \sys{Sanctuary} swap positions based on veracity metrics: \sys{Sanctuary} leads on accuracy (0.709 vs.\ 0.606) and macro-F1 (0.482 vs.\ 0.371).
In addition, \sys{Sanctuary} is a more generalisable system according to out-of-domain \averitec{} scores (\scifact{}: 0.490 vs.\ 0.390, \climatecheck{}: 0.460 vs.\ 0.000, \climatefever{}: 0.325 vs.\ 0.149), despite losing on the \averitec{} 2025 shared task. 

\begin{figure}[t]
  \includegraphics[width=\columnwidth]{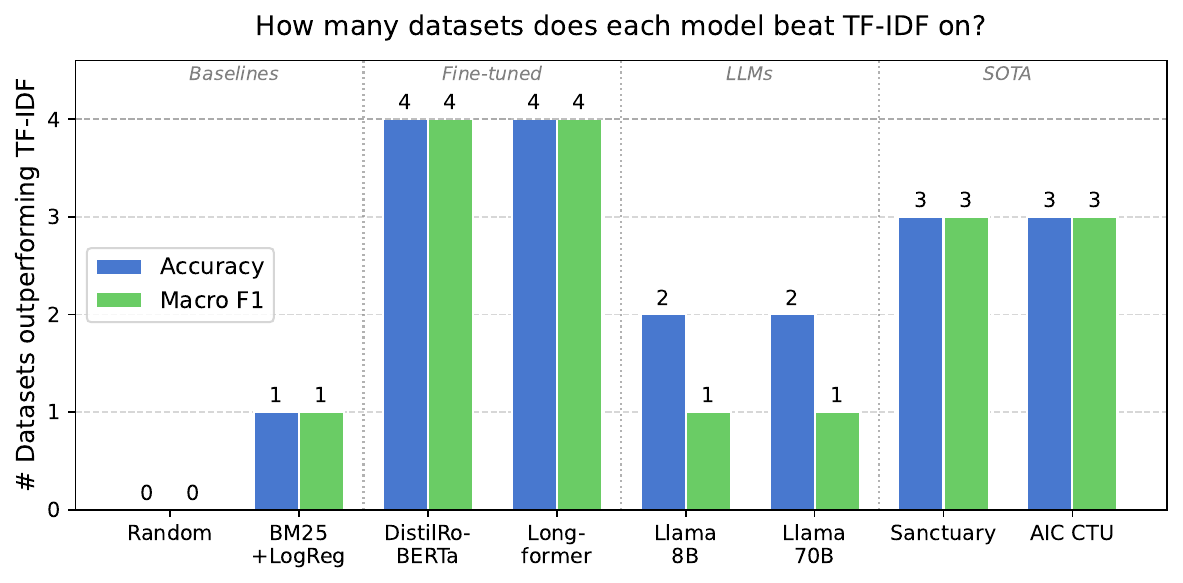}
  \caption{Number of datasets (out of 4) on which each model outperforms \sys{TF-IDF + LogReg}
    on accuracy (blue) and macro F1 (green).
    }
  \label{fig:beats-tfidf}
\end{figure}

\paragraph{Classical baselines reveal benchmark limitations.}
The dataset taxonomy (Figure~\ref{fig:taxonomy}) predicts where classical methods
will succeed: datasets in the \emph{web/Wikipedia evidence} row (\averitec{},
\climatefever{}) have claims and evidence drawn from the same register and vocabulary,
so lexical overlap is a reliable signal.
Datasets in the \emph{scientific evidence} row with informal claims (\climatecheck{})
have the largest vocabulary gap, suggesting the problem is retrieval coverage and not evidence understanding.
Only \scifact{} (scientific evidence, expert-written claims) rewards semantic reasoning
with a clear neural advantage.
Benchmark's position in Figure~\ref{fig:taxonomy} determines whether classical baselines are competitive, independently of model quality.
Figure~\ref{fig:beats-tfidf} shows that fine-tuned transformers are able to beat \sys{TF-IDF + LogReg} baseline on all datasets, while both strong evidence-conditioned systems and zero-shot LLMs fail on the climate and social-media corpora, both requiring domain transfer.

\paragraph{Retrieval quality is the primary bottleneck.}
Error analysis and oracle retrieval experiments (Table~\ref{tab:oracle}) indicate
that veracity performance is bounded by retrieval quality rather than model capacity.
When gold evidence is provided veracity accuracy improves, indicating that errors are caused predominantly by retrieval failures rather than veracity reasoning. Investing in better retrieval, particularly for large and informal corpora, yields greater gains than replacing the veracity model.
For highly domain-specific datasets, such as \climatecheck{} and \climatefever{}, domain-adaptive pretraining~\cite{gururangan-etal-2020-dont} and adaptive retrieval~\cite{asai2024self} could boost retrival performance, as seen in \climatecheck{} 2025 challenge~\cite{climatecheck2025}.
Based on our findings, we developed recommendations for further AFC evaluation in Appendix~\ref{app:recommendations}.

\section{Conclusion}
\label{sec:conclusion}

We present a cross-dataset benchmark study evaluating evidence retrieval and claim veracity prediction across \averitec{}, \scifact{}, \climatecheck{}, and \climatefever{}, covering a model range from sparse methods to \averitec{}~2025 winner systems.
Our findings demonstrate that simple baselines remain necessary lower bounds: claim-only \sys{TF-IDF + LogReg} outperform evidence-conditioned zero-shot LLMs and top-performing systems on \climatecheck{}, highliting that misleading retrieval can substantially degrade AFC systems. System rankings are strongly domain- and metric-dependent: single \sys{Sanctuary} spans 0.39 macro-F1 across datasets, and the \averitec{}~2025 winner and runner-up change rankings depending on the evaluation metric and dataset.
Retrieval remains the primary bottleneck: replacing retrieved evidence with gold annotations improves accuracy across models and datasets.

These findings show the need for cross-domain AFC evaluations with mandatory classical baselines and explicit retrieval-veracity decoupling. Until AFC systems are evaluated and prove their reliability across diverse domains, leaderboard rankings should not be taken as proof for real-world utility.

\section*{Limitations}

Our study covers four datasets, hence the conclusions may not generalise to all AFC domains or
claim types.
FEVER~\cite{fever} and FEVERous~\cite{feverous} were excluded due to their scale
(300K+ claims, millions of evidence documents), which exceeded our computational budget.
We do not evaluate multi-domain or multilingual AFC~\cite{petroni-etal-2021-kilt}, nor
recent generative AFC systems beyond \sys{Llama~3.1}~\cite{llama3modelcard}.
Baseline and fine-tuned transformer models are evaluated on claim text only (no retrieved evidence); oracle-retrieval results are reported in Table~\ref{tab:oracle} in the main text.
LLM experiments use a single prompt template per dataset; prompt sensitivity is not evaluated. Although we use two annotators to analyse failure modes, the agreement is moderate. Moreover, failure analysis covers only a subset of error cases and therefore could not be interpreted as dataset-wide estimate of annotation noise.

\section*{Ethical Considerations}

The datasets used in this study are publicly released research benchmarks
(\averitec{}, \scifact{}, \climatefever{}, \climatecheck{}), each with licences permitting academic use.
No new data were collected, and all claims and evidence documents are taken from the original datasets.
Our work evaluates existing models and datasets rather than deploying a
production fact-checking system.
All Llama~3.1 experiments use the publicly available model weights released by Meta under their community licence and accessed via the official Hugging Face repository. Our findings
identify measurement limitations in current benchmarks and do not endorse or refute any individual claim.

\section*{Generative AI Usage}
In this work, generative AI tools, such as ChatGPT~\cite{chatgpt}, were used to check for grammar mistakes and typos. The tool was used to enhance readability and the quality of the written text.

\bibliography{custom}

\appendix

\section{Evaluation Metrics}
\label{app:metrics}

\paragraph{Retrieval metrics.}
Let $E^*$ denote the set of gold-annotated relevant documents for a claim, and
$\hat{E}_K$ the set of top-$K$ documents returned by the retrieval system.
Recall@$K$ measures coverage of the gold set; Precision@$K$ measures result quality;
F1@$K$ is their harmonic mean:
\begin{align}
  \mathrm{R@}K &= \frac{|\hat{E}_K \cap E^*|}{|E^*|}, \qquad
  \mathrm{P@}K = \frac{|\hat{E}_K \cap E^*|}{K}, \label{eq:recall-prec}\\[4pt]
  \mathrm{F1@}K &= \frac{2\cdot\mathrm{P@}K\cdot\mathrm{R@}K}{\mathrm{P@}K+\mathrm{R@}K}. \label{eq:f1atk}
\end{align}
A document is counted as relevant if it is annotated as \emph{Supports} or \emph{Refutes}.
Recall@$K$ typically plateaus at $K{>}|E^*|$, which is why it is constant across $K\in\{5,10,20\}$
for \averitec{}, where the median gold set contains a single evidence document.

\paragraph{BM25 scoring.}
The Okapi BM25 score for document $d$ against query $q$ is~\cite{Robertson2009bm25}:
\begin{equation}
  \mathrm{BM25}(d,q) = \sum_{t \in q} \mathrm{IDF}(t)\cdot
    \frac{f_{td}\,(k_1{+}1)}{f_{td}+k_1\!\left(1{-}b{+}b\dfrac{|d|}{\overline{dl}}\right)},
  \label{eq:bm25}
\end{equation}
where $f_{td}$ is the term frequency of $t$ in $d$, $|d|/\overline{dl}$ is the
relative document length, and $k_1{=}1.5$, $b{=}0.75$ are standard smoothing parameters.
The length normalisation term $b\cdot|d|/\overline{dl}$ penalises long documents, which
explains why BM25 underperforms TF-IDF on long Wikipedia and scientific-abstract corpora.

\paragraph{Macro F1.}
For $C$ classes, macro-averaged F1 is:
\begin{equation}
  \text{Macro-F1} = \frac{1}{C}\sum_{c=1}^{C}\text{F1}_c, \label{eq:macrof1}
\end{equation}
where $\text{F1}_c = 2P_cR_c/(P_c{+}R_c)$ is the per-class F1.
Macro F1 weights each class equally, making it robust to label imbalance---unlike
accuracy, which is dominated by the majority class.
In our datasets, the majority class can constitute up to 60\% of samples,
so macro F1 and accuracy can differ by up to 20 points for the same model.

\paragraph{Why Recall@$K$/F1@$K$ rather than MRR.}
MRR rewards only the rank of the \emph{first} relevant document. Many claims in our datasets
require several gold documents before they become verifiable (\climatefever{}: 2--5 evidence
passages; \scifact{}: multiple rationale sentences), so the quantity that constrains downstream
veracity prediction is how much of the \emph{full} gold set is retrieved---exactly what
Recall@$K$ and F1@$K$ measure. For completeness, Table~\ref{tab:mrr-ndcg} reports MRR and
nDCG@10 for the sparse retrievers. The system ordering is unchanged: \sys{TF-IDF} $\geq$
\sys{BM25} on nDCG@10 on all three datasets, and the two are essentially tied on \scifact{}
MRR (0.1586 vs.\ 0.1590). No conclusion in Section~\ref{sec:res} therefore depends on the
choice of retrieval metric. \averitec{} is excluded because its evidence is annotated as
question--answer pairs and scored with Hungarian METEOR rather than document-level relevance.

\begin{table}[h]
\centering
\small
\setlength{\tabcolsep}{4pt}
\begin{tabular}{llcc}
\toprule
\textbf{Dataset} & \textbf{Retriever} & \textbf{MRR} & \textbf{nDCG@10} \\
\midrule
\multirow{3}{*}{\climatecheck{}}
  & \sys{TF-IDF} & \textbf{0.0707} & \textbf{0.0711} \\
  & \sys{BM25}   & 0.0586 & 0.0616 \\
  & \sys{Random} & 0.0000 & 0.0000 \\
\midrule
\multirow{3}{*}{\scifact{}}
  & \sys{TF-IDF} & 0.1586 & \textbf{0.2149} \\
  & \sys{BM25}   & \textbf{0.1590} & 0.2106 \\
  & \sys{Random} & 0.0011 & 0.0024 \\
\midrule
\multirow{3}{*}{\climatefever{}}
  & \sys{TF-IDF} & \textbf{0.2462} & \textbf{0.2388} \\
  & \sys{BM25}   & 0.2381 & 0.1926 \\
  & \sys{Random} & 0.0096 & 0.0029 \\
\bottomrule
\end{tabular}
\caption{MRR and nDCG@10 for sparse retrieval, reported alongside the Recall@$K$/F1@$K$ results
of Table~\ref{tab:retrieval_results}. Rank-based metrics give the same system ordering.}
\label{tab:mrr-ndcg}
\end{table}

\paragraph{AVeriTeC Hungarian METEOR.}
\averitec{} uses different retrieval evaluation system: evidence is annotated as
question--answer (QA) pairs, and a retrieved document is considered adequate if its
METEOR similarity~\cite{meteor} to any gold QA pair meets a threshold of 0.25.
Optimal assignment between retrieved documents and gold QA pairs is solved with the
Hungarian algorithm~\cite{hungarian}, yielding the AVeriTeC score used in the official
shared-task evaluation.

\section{Dataset Statistics and Characteristics}
\label{app:datasets}

\begin{table}[h]
\centering
\small
\setlength{\tabcolsep}{4pt}
\begin{tabular}{lrrrrc}
\toprule
\textbf{Dataset} & \textbf{Train} & \textbf{Dev} & \textbf{Test} & \textbf{Total} & \textbf{Imbal.} \\
\midrule
\averitec{}     & 4{,}626 & 578 & 579  & 5{,}783 & $\approx$5.0 \\
\scifact{}      & 1{,}127 & 141 & 141  & 1{,}409 & $\approx$1.2 \\
\climatecheck{} & 2{,}559 & 320 & 320  & 3{,}199 & $\approx$3.7 \\
\climatefever{} & 6{,}140 & 768 & 767  & 7{,}675 & $\approx$4.2 \\
\bottomrule
\end{tabular}
\caption{Dataset split sizes (80/10/10) and label imbalance ratio (most frequent /
least frequent class, estimated from per-class F1 of the random baseline).}
\label{tab:dataset-stats}
\end{table}

Table~\ref{tab:dataset-stats} reports split sizes and label imbalance ratios for all four
datasets

\subsection{Individual Dataset Descriptions and Data Quality}
\label{app:dataset-desc}

\paragraph{AVeriTeC.}
\averitec{}~\cite{averitec} contains 5,783 real-world political and
social media claims sourced from verified fact-checking websites.
Evidence is provided as human-annotated question--answer (QA) pairs instead of raw passages, making it a different retrieval protocol.
The corpus spans 32,818 web-scraped documents from diverse sources, resulting in high variance
in document length and style.

\paragraph{SciFact.}
\scifact{}~\cite{scifact} is a curated scientific fact-checking dataset
containing 1,409 expert-written claims about biomedical findings.
Evidence consists of 5,183 research abstracts from the Semantic Scholar Open Research Corpus
(S2ORC), and each claim is paired with one or more rationale sentences extracted from abstracts.
This dataset is the smallest of four and has the most balanced label distribution.

\paragraph{ClimateCheck.}
\climatecheck{}~\cite{climatecheck2025} focuses on climate misinformation drawn from social media posts.
Claims have informal language and are verified against 394,269 scientific
abstracts. The dataset has high label imbalance, with \emph{Refutes} constituting ~12\% of annotations.

\paragraph{ClimateFEVER.}
\climatefever{}~\cite{climatefever} extends the FEVER framework to climate
science, containing 7,675 claims and Wikipedia sentence-level passages as evidences.
Each claim is supported by up to five Wikipedia sentences, often from different articles. The dataset has overrepresented \emph{Supports} class.

\paragraph{Duplicate removal.}
We removed duplicate claims using exact-string matching on the normalised (lowercased,
whitespace-collapsed) claim text.
On \climatefever{}, this eliminated a small number of claims that appeared in both the
FEVER training set and the ClimateFEVER collection.

\paragraph{Noisy text.}
We detected and removed gibberish text using a combination of language identification
and perplexity-based filtering.\footnote{%
  We used \texttt{laurievb/OpenLID} for language detection and
  \texttt{alvations/langdetect} for perplexity filtering, both available on HuggingFace.}
For web-scraped \averitec{} documents, we additionally applied HTML tag stripping,
Unicode normalisation, and whitespace cleanup to recover readable text from raw scrapes.

\paragraph{Input text length.}
All models receive the full claim text as input.
For evidence, documents exceeding 4,096 tokens were truncated to fit within \sys{Longformer}'s context window; \sys{DistilRoBERTa} receives the first 512 tokens. Sparse retrieval methods (TF-IDF, BM25) and logistic regression classifiers operate on full-document term frequencies without truncation.




\subsection{Evidence Corpus Characteristics}
\label{app:corpus}

\begin{table}[h]
\centering
\small
\setlength{\tabcolsep}{4pt}
\begin{tabular}{lrrl}
\toprule
\textbf{Dataset} & \textbf{Corpus size} & \textbf{Avg.\ len.} & \textbf{Source} \\
\midrule
\averitec{}     & 32{,}818  & variable & Web pages \\
\scifact{}      & 5{,}183   & $\sim$250 tok. & S2ORC abst \\
\climatecheck{} & 394{,}269 & $\sim$200 tok. & Sci.\ abst \\
\climatefever{} & 5{,}240   & $\sim$80 tok.  & Wikipedia sent. \\
\bottomrule
\end{tabular}
\caption{Evidence corpus characteristics. Average length is estimated from
pre-processing statistics.}
\label{tab:corpus-stats}
\end{table}

Table~\ref{tab:corpus-stats} summarises the evidence corpus characteristics.
\averitec{}'s web-scraped corpus has the highest token variability: some documents could be single sentences, while others are multi-page articles.
Evidence documents exceeding 4,096 tokens were truncated when fed to \sys{Longformer}.
\scifact{} and \climatecheck{} use scientific abstracts, which fit within
transformer context window without truncation.
\climatefever{} uses sentence-level Wikipedia passages, which are the shortest evidences but require multi-sentence reasoning: gold evidence typically comprises 2-5 sentences from different passages.

\subsection{Label Distribution Visualisations}

\begin{figure*}[t]
  \includegraphics[width=\linewidth]{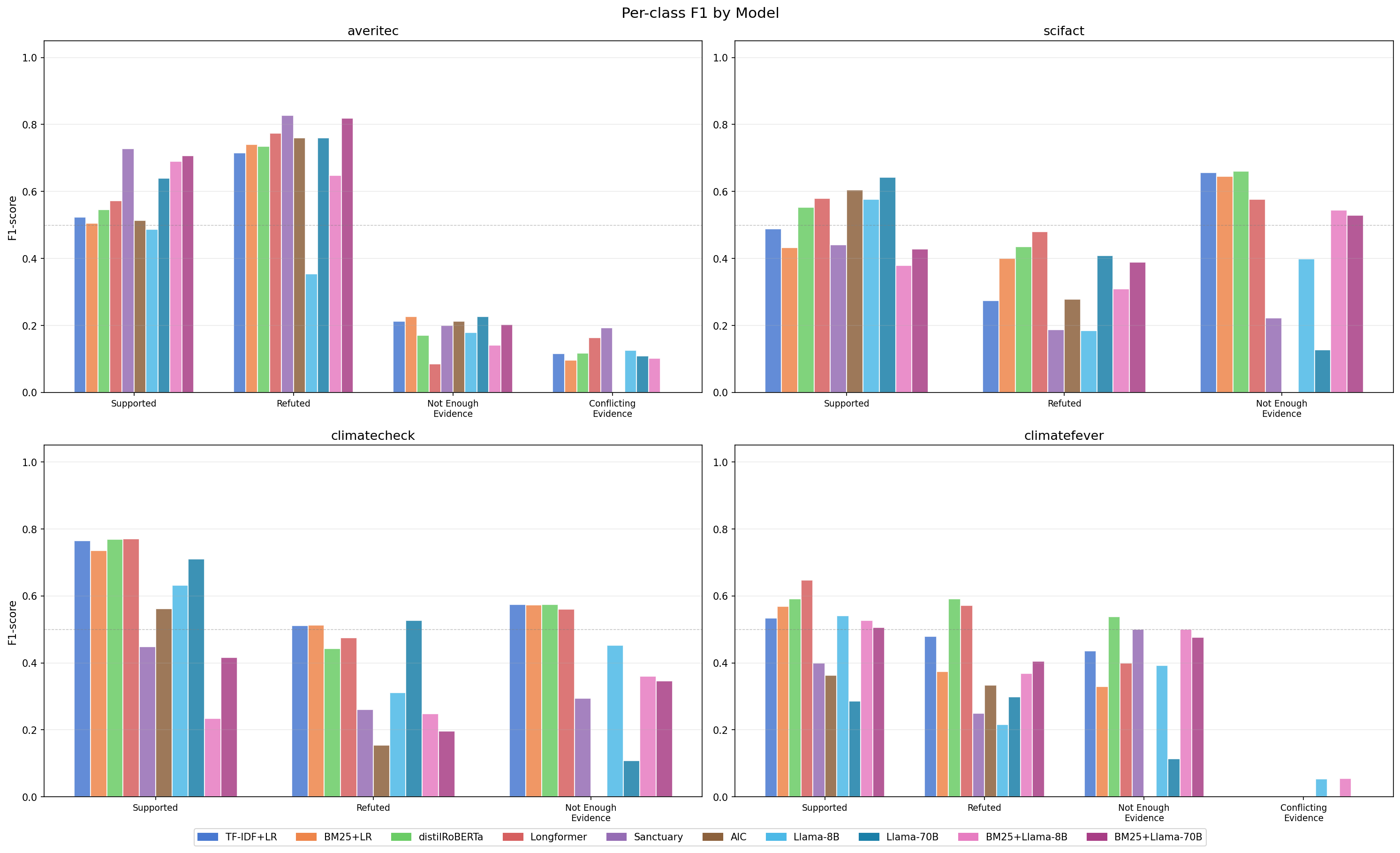}
  \caption{Per-class macro F1 across all models and datasets.
    Each bar shows the F1 for one label class; missing bars indicate zero F1.
    The rare \emph{Conflicting Evidence} class is effectively unpredictable by all models.
    On \scifact{} and \climatecheck{}, \emph{Refuted} is the hardest class,
    reflecting annotation sparsity.}
  \label{fig:per-class-f1}
\end{figure*}

Figure~\ref{fig:per-class-f1} and Table~\ref{tab:per_class} reveal a consistent pattern: models score substantially higher on dominant labels (e.g., \emph{Refutes} on \averitec{}, \emph{Supports} on \climatefever{}) and near zero on rare classes.
The \emph{Conflicting Evidence / Cherry-picking} label in \averitec{} and \climatefever{} is almost unpredictable by all systems, with most models scoring F1 $<$ 0.20 on this class.
On \climatecheck{}, \emph{NEI} class is learned more reliably (F1 $>$ 0.40 for transformer models) due to its large representation in the training data.

\begin{figure*}[t]
  \includegraphics[width=\linewidth]{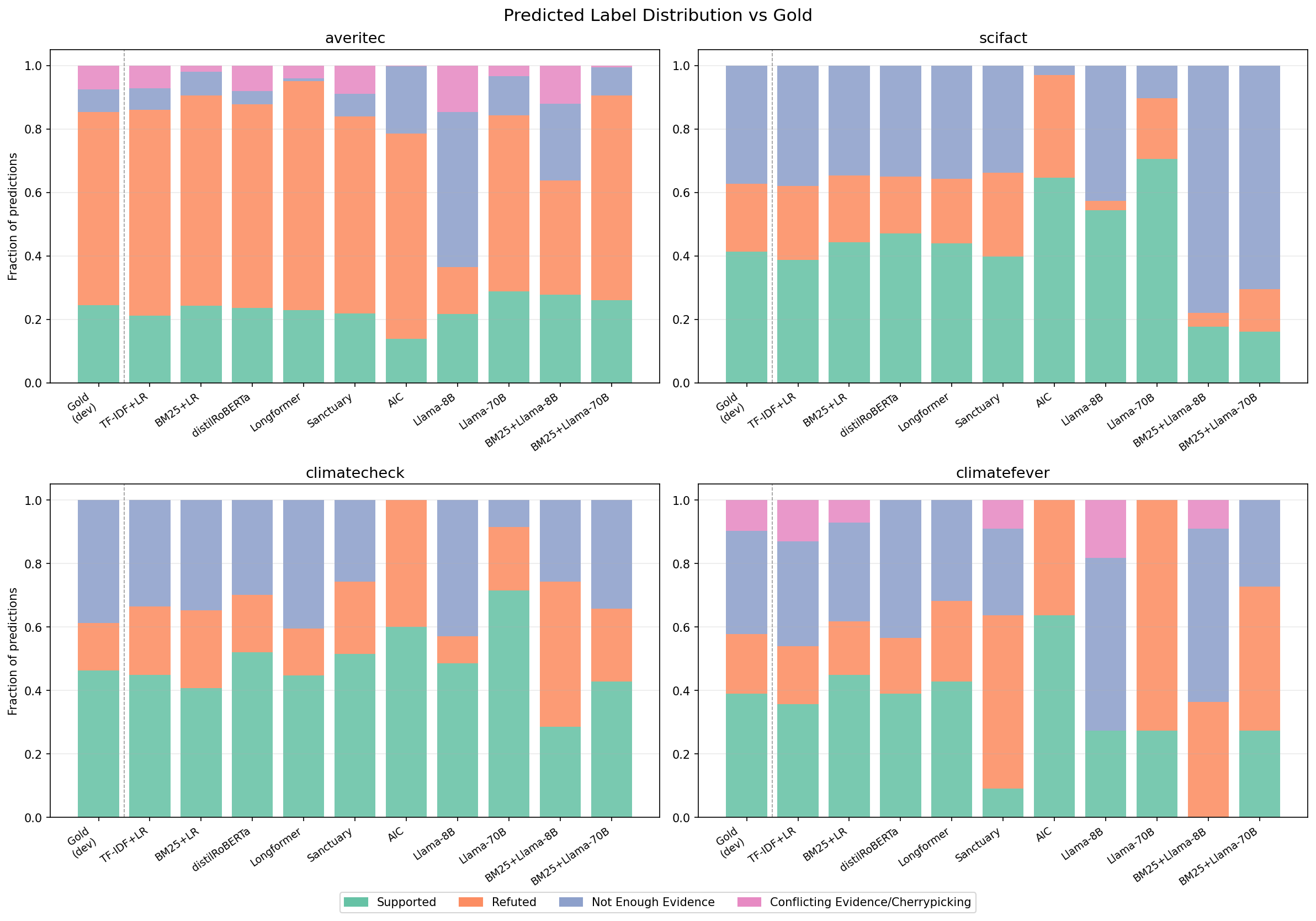}
  \caption{Predicted label distribution vs.\ gold label distribution.
    Each panel shows, for each model, the fraction of predictions assigned to each label.
    Models systematically over-predict the majority class, most severely on \averitec{}
    (\emph{Refuted}) and \climatefever{} (\emph{Supports}).}
  \label{fig:label-dist}
\end{figure*}

Figure~\ref{fig:label-dist} confirms that all models are
biased toward the dominant class.
On \averitec{}, models over-predict \emph{Refuted}, which accounts for nearly 60\% of
training labels, while several models assign \emph{Conflicting Evidence} to zero test claims.
On \climatefever{}, the mismatch between predicted and gold distributions is the most obvious: \sys{Llama~3.1-70B} collapses in predicting \emph{Refutes} for over 85\% of claims despite this label being $<$20\% of the gold data.
Unlike neural models, sparse methods are closer to training distribution.

\section{Experimental Configuration}
\label{app:supplementary}


All fine-tuning experiments were run on a single NVIDIA A100 (40~GB) GPU.
LLM inference (\sys{Llama 3.1-8B} and 70B) was performed on the same hardware using 4-bit NF4 quantisation for the 70B model to fit within GPU memory.


\subsection{Hyperparameter Summary}
\label{app:hyperparams}

Table~\ref{tab:hyperparams} summarises all hyperparameters used in fine-tuned model
experiments.
We performed no dataset-specific hyperparameter search; all settings follow the
configuration of \citet{calamai2025benchmarking} to enable fair comparison.

\begin{table}[h]
\centering
\small
\setlength{\tabcolsep}{4.5pt}
\begin{tabular}{lcc}
\toprule
\textbf{Hyperparameter} & \textbf{DistilRoBERTa} & \textbf{Longformer} \\
\midrule
Learning rate           & $5 \times 10^{-5}$     & $5 \times 10^{-5}$  \\
Warmup ratio            & 0.1                    & 0.1                 \\
Weight decay            & 0.01                   & 0.01                \\
Max epochs              & 10                     & 10                  \\
Early stopping metric   & Macro F1 (dev)         & Macro F1 (dev)      \\
Batch size              & 16                     & 4                   \\
Gradient accumulation   & 1                      & 4 steps             \\
Effective batch size    & 16                     & 16                  \\
Max input length        & 512 tokens             & 4{,}096 tokens      \\
Precision               & fp16                   & fp16                \\
Optimiser               & AdamW                  & AdamW               \\
Random seeds            & 26, 42, 123            & 26, 42, 123         \\
\bottomrule
\end{tabular}
\caption{Hyperparameters for fine-tuned transformer models.}
\label{tab:hyperparams}
\end{table}

\paragraph{Baselines.}
The \sys{Random} baseline uses scikit-learn's \texttt{DummyClassifier} with uniform
label sampling.
The \sys{TF-IDF + LogReg} and \sys{BM25 + LogReg} baselines use scikit-learn's
\texttt{TfidfVectorizer} (sublinear TF, no maximum feature limit), \texttt{BM25Okapi},
and \texttt{LogisticRegression} (C=1.0, class\_weight=\texttt{balanced},
max\_iter=1000) trained on the full training split.

\subsection{Oracle Retrieval}
\label{app:oracle}

Oracle-retrieval experiments are described in Section~\ref{sec:res} and results are
reported in Table~\ref{tab:oracle} in the main text.
Gold evidence was obtained from the original dataset annotations (supporting or refuting
passages labelled by human annotators).
For \averitec{}, the QA-pair evidence annotations were concatenated as a single evidence
string per claim.

\subsection{Significance Testing}
\label{app:significance}

We report 95\% bootstrap confidence intervals~\cite{efron1994introduction} using 1,000
resamples at the example level following best practices~\cite{dror-etal-2018-hitchhikers}.
A system is considered not significantly better than the \sys{TF-IDF} baseline if the
CIs of the two systems overlap.
Multi-seed experiments report mean and standard deviation across seeds.

\section{Zero-Shot Prompting Templates}
\label{app:prompts}

This section documents the prompt templates used for \sys{Llama~3.1-8B} and
\sys{Llama~3.1-70B} across all four datasets.
Templates share a common structure: a system instruction defining the task,
followed by the claim and evidence, and a chain-of-thought reasoning step before
the final verdict~\cite{wei2022chain}.

\paragraph{Template for 3-class datasets (\scifact{} and \climatecheck{}).}
Used for datasets with labels \emph{Supports}, \emph{Refutes}, \emph{Not Enough Information}.
\begin{quote}\small
\texttt{%
[SYSTEM] You are an expert fact-checking assistant.\\
Given a claim and a piece of evidence, determine the relationship.\\
\\
Claim: \{claim\}\\
Evidence: \{evidence\}\\
\\
Think step by step:\\
1. What does the evidence say?\\
2. Does it directly address the claim?\\
3. What is the logical relationship?\\
\\
Choose one verdict:\\
- Supports: evidence directly confirms the claim\\
- Refutes: evidence directly contradicts the claim\\
- Not Enough Information: evidence is insufficient to decide\\
\\
Reasoning: <your reasoning>\\
Verdict: <one of: Supports / Refutes / Not Enough Information>
}
\end{quote}

\paragraph{Template for 4-class datasets (\averitec{} and \climatefever{}).}
Extended with the \emph{Conflicting Evidence / Cherry-picking} label, following
the annotation guidelines of each dataset.
\begin{quote}\small
\texttt{%
[SYSTEM] You are an expert fact-checking assistant.\\
Given a claim and evidence, assign one of four labels.\\
\\
Claim: \{claim\}\\
Evidence: \{evidence\}\\
\\
Labels:\\
- Supports: evidence directly confirms the claim\\
- Refutes: evidence directly contradicts the claim\\
- Not Enough Evidence: evidence is insufficient to decide\\
- Conflicting Evidence: parts of the evidence both support and contradict the claim (cherry-picking)\\
\\
Think step by step before answering.\\
\\
Reasoning: <your reasoning>\\
Verdict: <one of: Supports / Refutes / Not Enough Information / Conflicting Evidence>
}
\end{quote}

Chain-of-thought guidance was enabled for all datasets after a 50-example validation
check showed that reasoning steps improved macro F1 on \averitec{} and \scifact{}.
On \climatefever{} the CoT step did not consistently improve performance and was
therefore ablated for that dataset; claim-only prompts (without evidence) were also
evaluated to quantify the retrieval contribution.

\section{Detailed Performance Analysis}
\label{app:analysis}

\subsection{Failure Mode Analysis}
\label{app:error-analysis}
We sampled 40 veracity prediction failure modes. Two independent LLM annotators, Claude Opus 4.8~\footnote{\url{https://platform.claude.com/docs/en/models/opus-4-8/overview}} and Claude Sonet 4.6~\footnote{\url{https://platform.claude.com/docs/en/models/sonnet-4-6/overview}}, performed error labeleing according to \citet{calamai2025benchmarking} classification:
(1)~a \textbf{model error} (the label is unambiguous and the model is wrong);
(2)~an \textbf{annotation mistake} (the gold label appears incorrect on inspection); or
(3)~a \textbf{debatable errors} (the label depends on an interpretation not fixed by the guidelines).
Table~\ref{tab:error_examples} gives representative examples of each type across datasets.

Both LLM annotators worked independently from the same written three-way guideline, using the class definitions above. On the $n{=}40$ sampled cases, inter-annotator agreement is Cohen's $\kappa = 0.55$ with a raw agreement of $0.72$, indicating only moderate agreement and confirming how ambiguous these cases are. Both annotators marked 38\% of the sampled ``errors'' as annotation mistakes or debatable cases rather than genuine model errors. Because that fraction is large, score gaps below ${\sim}5$ accuracy points on these datasets can be explained by label noise alone.

\paragraph{Retrieval failures: corpus coverage and vocabulary mismatch.}
The dominant failure mode on \climatecheck{} is vocabulary mismatch:
informal claim language (slang, hashtags, colloquial references to weather events)
shares few keywords with formal scientific articles.
On \climatefever{}, failures cluster on claims that require multi-passage reasoning:
no single Wikipedia passage is sufficient to support or refute the claim,
and the system fails to retrieve full set of evidential passages.
On \scifact{}, retrieval failures occur primarily for claims using synonymous terminology not present in the abstract.
On \averitec{}, the QA-style evidence structure means that relevant evidence is often
fragmented across multiple retrieved documents.

\paragraph{Veracity errors: label ambiguity.}
The boundary between \emph{Refutes} and \emph{NEI} is blurred across all datasets.
Evidence that partially contradicts a claim may be labelled either category depending on
how strictly the annotator interprets ``sufficient contradiction.''
Approximately 30\% of sampled \climatefever{} errors involve the \emph{NEI} class,
with the model predicting \emph{Refutes} or vice versa.
Supporting the finding of \citet{calamai2025benchmarking} that \emph{NEI} generates the most
disagreement in climate NLP benchmarks.
On \climatecheck{}, social-media claims tend to express the same scientific content in varied forms, leading models to predict \emph{NEI} when the claim is technically supported at a semantic level but not lexically.
On \averitec{}, the four-label scheme (including \emph{Conflicting Evidence}) introduces additional ambiguity, and models systematically over-predict \emph{Refutes}.

\paragraph{Domain mismatch in \averitec{} 2025 winning systems.}
\sys{Sanctuary} and \sys{AIC CTU} were developed for FEVER-style encyclopaedic claims.
On \climatecheck{}, their veracity components underperform domain-fine-tuned \sys{Longformer}.
Error inspection reveals that these systems frequently misclassify informal paraphrases of
scientific findings as \emph{NEI}, apparently because the informal phrasing is not recognised as
entailing or contradicting the formal scientific evidence.
Zero-shot LLMs face a compounding problem: they receive informal social-media claims paired with formal scientific evidence, and without domain fine-tuning they cannot reliably recognise entailment across such a wide vocabulary gap.
This failure mode does not appear on \scifact{} and \averitec{}, where claim language is closer evidence corpus language.

\paragraph{Evidence length bias in retrieval.}
On \scifact{} and \climatecheck{}, longer evidence documents are retrieved more often
because they contain more claim-adjacent tokens, inflating Recall@$K$ independently of
true relevance.
This creates a superficial length bias most pronounced on \climatecheck{},
where scientific articles span many paragraphs.




\begin{table*}[t]
\centering
\small
\setlength{\tabcolsep}{4pt}
\renewcommand{\arraystretch}{1.2}
\begin{tabular}{p{2cm} p{5.2cm} p{4.5cm} p{1.6cm} p{1.6cm}}
\toprule
\textbf{Dataset} & \textbf{Claim (paraphrased)} & \textbf{Retrieval / evidence note}
  & \textbf{Gold} & \textbf{Pred.\ / type} \\
\midrule
\averitec{}
  & ``A vaccine candidate reduced symptomatic COVID-19 by 90\%''
  & Retrieved web snippet reports a different trial arm (70\%);
    relevant QA pair is fragmented across two documents
  & Supported & Refuted\newline {\scriptsize (genuine)} \\
\averitec{}
  & ``Company X lobbied against safety regulation''
  & Gold label is Conflicting Evidence but only one side is present
    in the evidence; annotation schema conflates two decisions
  & Conflicting & Refuted\newline {\scriptsize (annotation)} \\
\scifact{}
  & ``Drug Y inhibits tumour growth via pathway Z''
  & Abstract uses synonym ``suppresses'' for ``inhibits'';
    TF-IDF retrieval misses the rationale sentence
  & Supports & NEI\newline {\scriptsize (genuine)} \\
\scifact{}
  & ``Protein X regulates inflammation''
  & Abstract states X \emph{modulates} inflammation;
    whether modulation counts as regulation is ambiguous
  & Supports & NEI\newline {\scriptsize (debatable)} \\
\climatecheck{}
  & ``LOL the ice caps are disappearing fast \#climatecrisis''
  & Social-media phrasing shares no tokens with scientific abstract;
    model retrieves unrelated document about sea-surface temperature
  & Supports & NEI\newline {\scriptsize (genuine)} \\
\climatecheck{}
  & ``Scientists proved global warming is fake news''
  & Ironic/sarcastic post; gold label treats it at face value
    (Refuted) though the author's intent is pro-science
  & Refuted & Supports\newline {\scriptsize (debatable)} \\
\climatefever{}
  & ``The Arctic is warming twice as fast as the global average''
  & Gold evidence requires combining two Wikipedia passages;
    single-passage retrieval returns only one fragment
  & Supports & NEI\newline {\scriptsize (genuine)} \\
\climatefever{}
  & ``CO$_2$ levels have been higher in pre-industrial eras''
  & Wikipedia passage is factually correct but claim's implicit
    implication (current warming is natural) is not addressed;
    boundary between NEI and Refuted is undefined by guidelines
  & NEI & Refuted\newline {\scriptsize (debatable)} \\
\bottomrule
\end{tabular}
\caption{Representative error examples per dataset.
  ``Genuine'' = clear model error; ``annotation'' = questionable gold label;
  ``debatable'' = guideline ambiguity.
  Claims are paraphrased for anonymity; evidence notes summarise the key failure mode.}
\label{tab:error_examples}
\end{table*}

\subsection{Per-Dataset Analysis}

\paragraph{\averitec{}.}
The dominant failure mode is evidence fragmentation: relevant information is often
distributed across multiple QA pairs in the annotation, but retrieval returns
fragments that are individually insufficient to support or refute the claim.
\sys{Sanctuary} systematically over-predicts \emph{Refuted} on \averitec{}
(Table~\ref{tab:per_class}), which partly reflects the high proportion of \emph{Refuted} claims in the training data and partly the fact that partially retrieved evidence often appears contradictory.
The \emph{Conflicting Evidence / Cherry-picking} label is almost unpredictable (F1 $<$ 0.18 for all models): annotation instructions mix two distinct annotation decisions
(contradictory evidence vs.\ selective use of one-sided evidence), making consistent annotation difficult.
Among the sampled errors, debatable cases concentrate on \emph{Conflicting} vs.\
\emph{Refuted} boundary (when a claim is factually wrong but the evidence also shows partial support).

\paragraph{\scifact{}.}
Failures cluster into two groups.
First, \textbf{lexical mismatch}: biomedical claims use precise terminology, but
the rationale sentence in the abstract uses synonymous or related vocabulary
(e.g., ``suppresses'' / ``inhibits''; ``modulates'' / ``regulates'').
TF-IDF and dense retrievers miss these
when the overlap is low, leading to \emph{NEI} predictions for claims that are actually supported by the retrieved abstract.
Second, \textbf{sentence-level granularity}: \scifact{} labels are grounded in specific rationale sentences, but if retrieval surfaces the abstract without the precise sentence, the veracity model receives insufficient information and chooses \emph{NEI}.
This is why oracle retrieval experiments yielded large accuracy improvements on \scifact{} ($+$18--19~pp for LLMs).
Annotation issues are relatively rare on \scifact{}, the expert-constructed claims and controlled vocabulary lead to consistent labels.

\paragraph{\climatecheck{}.}
This dataset presents the widest vocabulary gap: social-media claims
use slang, hashtags, abbreviations, and colloquial references to weather events, while the evidence corpus consists of formal scientific abstracts.
The failure mode is usually that
retrieval returns irrelevant abstracts because the claim text shares no tokens with any relevant document.
The relatively smaller oracle gain on \climatecheck{} ($+$14--15~pp for LLMs) reflects that even gold evidence provides limited information for claims expressed.
\sys{Sanctuary} and \sys{AIC CTU} fail substantially on this dataset because they were trained on encyclopaedic fact-checking, where claim and evidence share formal language.
A notable annotation issue is irony and sarcasm: social-media posts that mock climate denial are labeled \emph{Refuted} (the factual content of the ironic claim is false), since the model's failure to detect sarcasm.

\paragraph{\climatefever{}.}
Two failure modes dominate.
First, \textbf{multi-passage reasoning}: gold evidence for \climatefever{} claims
typically comprises 2--5 Wikipedia passages. A single-passage retrieval, even if it is relevant, is insufficient to resolve the claim.
Hence, largest oracle gains in the study ($+$20--22~pp).
Second, \textbf{NEI/Refutes boundary ambiguity}: approximately 30\% of sampled \climatefever{} errors involve the \emph{NEI} class, with the model predicting \emph{Refutes} or vice versa.
Once again, \emph{NEI} causes the most annotator disagreement in climate NLP benchmarks.
Additionally, the annotation instructions do not define whether ``Supports'' requires the evidence to \emph{entail} the claim or merely be \emph{consistent} with it.
\sys{Llama~3.1-70B} predicted \emph{Refutes} for over 85\% cases in  \climatefever{}, which shows that model has in-context tendency to treat any climate-related claim as false.

\paragraph{Cross-dataset patterns.}
We observe three across all datasets:
(1)~The \emph{Conflicting Evidence} label is universally hard: models score 0.00--0.20, suggesting the schema itself is
underspecified.
(2)~NEI over-prediction is the dominant error mode for LLMs on informal or
scientific-domain datasets, where retrieved evidence is often tangentially related
but insufficient for a confident verdict.
(3)~Annotation issues account for most failure cases on all datasets,
suggesting that performance margins below 5 accuracy points may not reliably distinguish model capability from label noise.

\begin{table*}[t]
\centering
\small
\begin{tabular}{ll cccc}
\toprule
  Dataset & Method & Supported & Refuted & NEI & Conflicting\\
\midrule
\multirow{10}{*}{{\averitec{}}} &
  \sys{Random} 
    & 0.2468	& 0.3835	& 0.1287	& 0.1395	\\
  & \sys{TF-IDF + LogReg} 
    & 0.3596	& 0.6995	& 0.2609	& 0.1622\\
  & \sys{BM25 + LogReg} 
    & 0.3868	& 0.6907	& 0.1667	& 0.0417\\
  & \sys{DistilRoBERTa} 
    & 0.4282	& 0.6971	& 0.1502	& 0.2199\\
  & \sys{Longformer} 
    & 0.4532	& 0.7121	& 0.1535	& 0.1920\\
  &\sys{Llama 8B} 
    & 0.4870	& 0.3536	& 0.1786	& 0.1261\\
  & \sys{Llama 70B} 
    & 0.6391	& 0.7595	& 0.2268	& 0.1091\\
  & \sys{BM25 + Llama 8B} 
    & 0.6897	& 0.6474	& 0.1410	& 0.1020\\
  & \sys{BM25 + Llama 70B} 
    & 0.7063	& 0.8185	& 0.2025	& 0.0000\\
  & \sys{Sanctuary} 
    & 0.7306	& 0.8271	& 0.1897	& 0.1795 \\
  & \sys{AIC CTU} 
    & 0.5131	& 0.7599	& 0.2128	& 0.0000\\
\midrule

\multirow{10}{*}{{\scifact{}}} &
  \sys{Random} 
    & 0.3508	& 0.3055	& 0.3578 & {-}\\
  & \sys{TF-IDF + LogReg} 
    & 0.3667	& 0.1940	& 0.6283 & {-}\\
  & \sys{BM25 + LogReg} 
    & 0.4358	& 0.2047	& 0.6296 & {-}\\
  & \sys{DistilRoBERTa} 
    & 0.4719	& 0.2445	& 0.5571 & {-}\\
  & \sys{Longformer} 
    & 0.4657	& 0.2903	& 0.5584 & {-}\\
  &\sys{Llama 8B} 
    & 	0.5772	& 0.1842	& 0.3982 & {-}\\
  & \sys{Llama 70B} 
    & 	0.6422	& 0.4091	& 0.1277 & {-}\\
  & \sys{BM25 + Llama 8B} 
    & 0.3793	& 0.3095	& 0.5439 & {-}\\
  & \sys{BM25 + Llama 70B} 
    & 0.4277	& 0.3894	& 0.5287 & {-}\\
  & \sys{Sanctuary} 
    & 0.7471	& 0.6810	& 0.6720 & {-} \\
  & \sys{AIC CTU} 
    & 0.7381	& 0.5273	& 0.1811 & {-}\\

\midrule
\multirow{10}{*}{{\climatecheck{}}} &
  \sys{Random} 
    & 0.4409 & 0.1654 & 0.3829 & {-}\\
  & \sys{TF-IDF + LogReg} 
    & 0.7008 & 0.6099 & 0.5375 & {-}\\
  & \sys{BM25 + LogReg} 
    & 0.6649 & 0.5455 & 0.5061 & {-}\\
  & \sys{DistilRoBERTa} 
    & 0.7257 & 0.6177 & 0.5323 & {-}\\
  & \sys{Longformer} 
    & 0.7212 & 0.6083 & 	0.5396 & {-}\\
  &\sys{Llama 8B} 
    & 0.6427 & 0.2973 & 0.4820 & {-}\\
  & \sys{Llama 70B} 
    & 0.7160 & 0.5035	 & 0.1244 & {-}\\
  & \sys{BM25 + Llama 8B} 
    & 0.3333 & 0.2576 & 	0.3959 & {-}\\
  & \sys{BM25 + Llama 70B} 
    & 0.6150 & 0.4667 & 0.4161 & {-}\\
  & \sys{Sanctuary} 
    & 0.7081 & 	0.4518 & 0.3709 & {-} \\
  & \sys{AIC CTU} 
    & 0.7106 & 0.3842 & 	0.1396 & {-}\\
\midrule
\multirow{10}{*}{{\climatefever{}}} &
  \sys{Random} 
    & 0.3041	& 0.2403	& 0.2619	& 0.129\\
  & \sys{TF-IDF + LogReg} 
    & 0.4348	& 0.4912	& 0.3762	& 0.1714\\
  & \sys{BM25 + LogReg} 
    & 0.5426	& 0.4000	& 0.4490	& 0.153\\
  & \sys{DistilRoBERTa} 
    & 0.5729	& 0.4935	& 0.4460	& 0.0238\\
  & \sys{Longformer} 
    & 0.5818	& 0.4101	& 0.4429	& 0.0000\\
  &\sys{Llama 8B} 
    & 0.5410	& 0.2162	& 0.3929	& 0.0541\\
  & \sys{Llama 70B} 
    & 0.2857	& 0.2993	& 0.1132	& 0.0000\\
  & \sys{BM25 + Llama 8B} 
    & 0.5263	& 0.3692	& 0.5000	& 0.0556\\
  & \sys{BM25 + Llama 70B} 
    & 0.5055	& 0.4054	& 0.4762	& 0.0000\\
  & \sys{Sanctuary} 
    & 0.4411	& 0.5108	& 0.5180	& 0.1278 \\
  & \sys{AIC CTU} 
    & 0.6133	& 0.4427	& 0.2259	& 0.0584\\
\bottomrule
\end{tabular}
\caption{
  Macro-F1 per-class breakdown.
}
\label{tab:per_class}
\end{table*}



\clearpage
\section{Frontier-LLM Verifier Experiments}
\label{app:frontier}

This appendix reports two additional experiments that hold the verifier fixed and vary only how
evidence is obtained. Both use a frontier closed-weight verifier, \sys{Claude Opus~4.8}, on fixed
samples of \averitec{} ($n{=}63$) and \climatecheck{} ($n{=}55$); the verifier never sees gold
labels.

\subsection{Three Evidence Conditions under an Identical Prompt}
\label{app:evidence-quality}

We run the same model with the same prompt under three evidence conditions:
(i)~\emph{claim-only}; (ii)~\emph{$+$ TF-IDF retrieved} evidence, the pipeline setting used
throughout the paper; and (iii)~\emph{$+$ gold} evidence, i.e.\ oracle retrieval
(Appendix~\ref{app:oracle}). The prompt is identical in all three conditions, so only the
evidence quality changes.

\begin{table}[htbp]
\centering
\small
\setlength{\tabcolsep}{4pt}
\begin{tabular}{lcc}
\toprule
\textbf{Evidence condition} & \textbf{\averitec{}} & \textbf{\climatecheck{}} \\
\midrule
claim-only            & 81.0\,/\,69.0 & 69.1\,/\,60.3 \\
$+$ TF-IDF retrieved  & 38.1\,/\,36.6 & 60.0\,/\,59.6 \\
$+$ gold (oracle)     & \textbf{87.3}\,/\,\textbf{71.2} & \textbf{70.9}\,/\,\textbf{63.1} \\
\bottomrule
\end{tabular}
\caption{\sys{Claude Opus~4.8} under three evidence conditions with an identical prompt
(accuracy\,/\,macro-F1), on fixed samples of \averitec{} ($n{=}63$) and \climatecheck{}
($n{=}55$). Only evidence quality varies across rows.}
\label{tab:claude-evidence}
\end{table}

The model is best with gold evidence (87.3 accuracy on \averitec{}), worst with retrieved
evidence (38.1), and in between with claim-only input (81.0). A difference of 49 points is
therefore driven by evidence quality alone. Since the model is fixed, this gap cannot come from
reasoning ability: retrieval is the bottleneck, and this holds even for a frontier model.

\subsection{Iterative Agentic Verification}
\label{app:agentic}

Agentic systems interleave retrieval and verification instead of retrieving
once~\cite{xie-etal-2025-fire,braun2025defame}. We test a \sys{FIRE}-style~\cite{xie-etal-2025-fire}
agent that retrieves and verifies in rounds, stopping early when confident: in round~1 the
verifier either commits to a veracity label or emits a \textsc{search} action, and only the
claims that request search receive a second retrieval round (top-8 passages, against 3 in the
one-shot setting) before being re-verified. Only 16/63 (25.4\%) of \averitec{} claims and
17/55 (30.9\%) of \climatecheck{} claims need a second round.

\begin{table}[htbp]
\centering
\small
\setlength{\tabcolsep}{4pt}
\begin{tabular}{lcc}
\toprule
\textbf{Verification strategy} & \textbf{\averitec{}} & \textbf{\climatecheck{}} \\
\midrule
one-shot TF-IDF retrieval  & 38.1\,/\,36.6 & 60.0\,/\,59.6 \\
\sys{FIRE}-style iterative & 47.6\,/\,43.2 & 56.4\,/\,55.5 \\
gold evidence (oracle)     & \textbf{87.3}\,/\,\textbf{71.2} & \textbf{70.9}\,/\,\textbf{63.1} \\
\bottomrule
\end{tabular}
\caption{Iterative retrieval-and-verification (accuracy\,/\,macro-F1) against one-shot retrieval
and oracle retrieval, same model and samples as Table~\ref{tab:claude-evidence}.
Second retrieval rounds are triggered for 25.4\% (\averitec{}) and 30.9\% (\climatecheck{}) of claims.}
\label{tab:fire}
\end{table}

The agent improves over one-shot retrieval on \averitec{} (47.6 vs.\ 38.1 accuracy) but not on
\climatecheck{} (56.4 vs.\ 60.0), and stays far below gold evidence (87.3 and 70.9). Even an
agentic method is therefore capped by retrieval quality rather than by the verifier.

\section{Recommendations for AFC Evaluation Design}
\label{app:recommendations}

Many fact-checking evaluations draw misleading conclusions because simple baselines are absent
and cross-dataset comparison is neglected---a concern echoed across NLP
benchmarking~\cite{guo2022survey,thakur2021beir,calamai2025benchmarking}.
We expand the five recommendations below.

\textbf{Always include classical sparse baselines.}
\sys{TF-IDF} and \sys{BM25} with logistic regression should be mandatory starting points
in any fact-checking evaluation.
Without these lower bounds, it is impossible to assess dataset difficulty or to measure beyond surface-level pattern matching.
As a practical threshold: if a proposed system fails to beat \sys{TF-IDF + LogReg} by
more than 5 accuracy points on an in-domain dataset, the claimed improvement may not be
reliable given typical annotation noise levels.

\textbf{Evaluate across multiple domains.}
Single-benchmark results are insufficient evidence of general capability.
Our results show that a system leading on one domain may perform at or below the sparse baseline on another, a 0.39 macro-F1 span for \sys{Sanctuary} across our four datasets.
Cross-dataset evaluation should include at least one dataset outside the system's training distribution; the dataset taxonomy in Figure~\ref{fig:taxonomy} provides a way to identify structurally different test conditions.

\textbf{Decouple retrieval and veracity evaluation.}
Current evaluation frameworks conflate retrieval and veracity errors, making it impossible
to know where to invest modelling effort.
We recommend oracle-retrieval experiments as a standard component: running veracity models
on gold evidence reveals the upper bound that better retrieval could achieve.

\textbf{Report per-label F1 alongside aggregate metrics.}
Macro F1 and accuracy can differ by up to 20 points under realistic class imbalance.
Reporting per-label F1 reveals whether a system genuinely predicts all label classes or
simply reproduces majority-class predictions.
This is especially important for rare labels such as \emph{Conflicting Evidence}, which our
experiments show to be effectively unpredictable (F1 $<$ 0.20) by all current systems.

\textbf{Quantify annotation quality.}
When model performance differences are small (often $<$5 accuracy points on our datasets),
annotation noise can explain the gap.
We recommend estimating annotation error rates through inter-annotator agreement or manual
sampling, and establishing a minimum reliable margin before claiming a system improvement.
Our error analysis found annotation issues in all four datasets; on \climatefever{}, a
Cohen's $\kappa = 0.334$ for evidence annotations~\cite{calamai2025benchmarking} suggests
that differences below $\sim$5 points are within the noise floor.

\end{document}